%% file: latex/acl_latex.tex
\pdfoutput=1

\documentclass[11pt]{article}

\usepackage[final]{acl}

\usepackage{times}
\usepackage{latexsym}
\usepackage{array}
\usepackage{booktabs} 
\usepackage{multirow}
\usepackage{tabularx}
\usepackage{graphicx}
\usepackage{makecell}
\usepackage{caption}
\usepackage{subcaption}
\usepackage[titletoc]{appendix}
\usepackage{textcomp}
\usepackage{xspace}
\usepackage{soul} 
\usepackage{color}
\usepackage{xcolor}
\usepackage{amsmath}
\usepackage{amsfonts}
\usepackage{mathrsfs}
\usepackage{bm}
\definecolor{shadecolor}{rgb}{0.92,0.92,0.92}  
\usepackage{tcolorbox} 
\usepackage[export]{adjustbox}
\definecolor{mygray}{RGB}{230,230,240}
\definecolor{myblue}{RGB}{175, 238, 235}
\definecolor{deepgreen}{rgb}{0.0, 0.5, 0.0}
\usepackage[linesnumbered,ruled,vlined]{algorithm2e}
\usepackage{multicol}
\usepackage{tikz}
\usetikzlibrary{shapes, positioning}
\usepackage{setspace}
\usepackage{hyperref}

\usepackage{enumitem}
\usepackage[skip=5pt]{caption}
\usepackage{xcolor} 
\usepackage{soul} 
\sethlcolor{white}
\soulregister\cite7 
\soulregister\citep7 
\soulregister\citet7 
\soulregister\ref7 
\soulregister\pageref7 
\soulregister\eqref7 

\newcommand{\ourtitle}{AMoPO}

\usepackage[T1]{fontenc}

\usepackage[utf8]{inputenc}

\usepackage{microtype}

\usepackage{inconsolata}

\usepackage{graphicx}
\usepackage{hyperref}
\title{AMoPO: Adaptive Multi-objective Preference Optimization without Reward Models and Reference Models}

\makeatletter
\renewcommand\@fnsymbol[1]{\ifcase#1\or $\dagger$ \or * \fi}
\makeatother
\author{
    \textbf{Qi Liu}\textsuperscript{1},
    \textbf{Jingqing Ruan}\thanks{Lead this project. \quad * Corresponding authors.}\textsuperscript{2},
    \textbf{Hao Li}\textsuperscript{2},
    \textbf{Haodong Zhao}\textsuperscript{2},
    \textbf{Desheng Wang}\textsuperscript{2},
    \textbf{Jiansong Chen}\textsuperscript{*2}, \\
    \textbf{Guanglu Wan}\textsuperscript{2},
    \textbf{Xunliang Cai}\textsuperscript{2},
    \textbf{Zhi Zheng}\textsuperscript{1},
    \textbf{Tong Xu}\textsuperscript{*1} \\
    \textsuperscript{1} University of Science and Technology of China, State Key Laboratory of Cognitive Intelligence, Hefei, China \\
    \textsuperscript{2} Meituan \\
    \texttt{\{ruanjingqing,lihao138,zhaohaodong04,wangdesheng08,chenjiansong\}@meituan.com} \\
    \texttt{liuqilq@mail.ustc.edu.cn} \\
    \texttt{\{zhengzhi97, tongxu\}@ustc.edu.cn} 
}

\begin{document}
\maketitle




\input{latex/sections_article/0_abs_lq}

\section{Introduction}

\input{latex/sections_article/1_intro_rjq}

\section{Related Work}

\input{latex/sections_article/2_related_rjq}

\section{Preliminary}

\input{latex/sections_article/3_pre_rjq}

\section{Methodology}


\input{latex/sections_article/4_RMoPO_rjq}

\section{Experiments}

\input{latex/sections_article/5_Experiment}

\vspace{-0.2cm}
\section{Conclusions}

\input{latex/sections_article/6_con}

\section*{Acknowledgments}

\input{latex/sections_article/9_ack}

\clearpage
\section*{Limitations}

\input{latex/sections_article/7_limitations}


\clearpage

\bibliography{latex/custom}

\clearpage

\appendix

\input{latex/sections_article/8_app}

\end{document}

%% file: latex/sections_article/0_abs_lq.tex
\begin{abstract}

\hl{Existing multi-objective preference alignment methods for large language models (LLMs) face limitations: (1) the inability to effectively balance various preference dimensions, and (2) reliance on auxiliary reward/reference models introduces computational complexity. 
To address these challenges, we propose Adaptive Multi-objective Preference Optimization (AMoPO), a novel framework that achieves dynamic balance across preference dimensions.
By introducing the multi-objective optimization paradigm to use the dimension-aware generation metrics as implicit rewards, AMoPO aligns LLMs with diverse preferences without additional reward models or reference models. 
We introduce an adaptive weight assignment mechanism that models the generation space as a Gaussian distribution, allowing dynamic prioritization of preference dimensions. 
Empirical results demonstrate that {~\ourtitle} outperforms state-of-the-art baselines by 28.5$\%$, and the experiments on 7B, 14B, and 32B models reveal the scaling ability of \ourtitle.}
Moreover, additional analysis of multiple dimensions verifies its adaptability and effectiveness.
\hl{These findings validate AMoPO's capability to achieve dimension-aware preference alignment, highlighting its superiority.}
Our codes and datasets are available at 
\href{https://github.com/Javkonline/AMoPO}{https://github.com/Javkonline/AMoPO}.

\end{abstract}

%% file: latex/sections_article/1_intro_rjq.tex

\begin{figure}[t]
    \centering
    \includegraphics[width=1.0\linewidth]{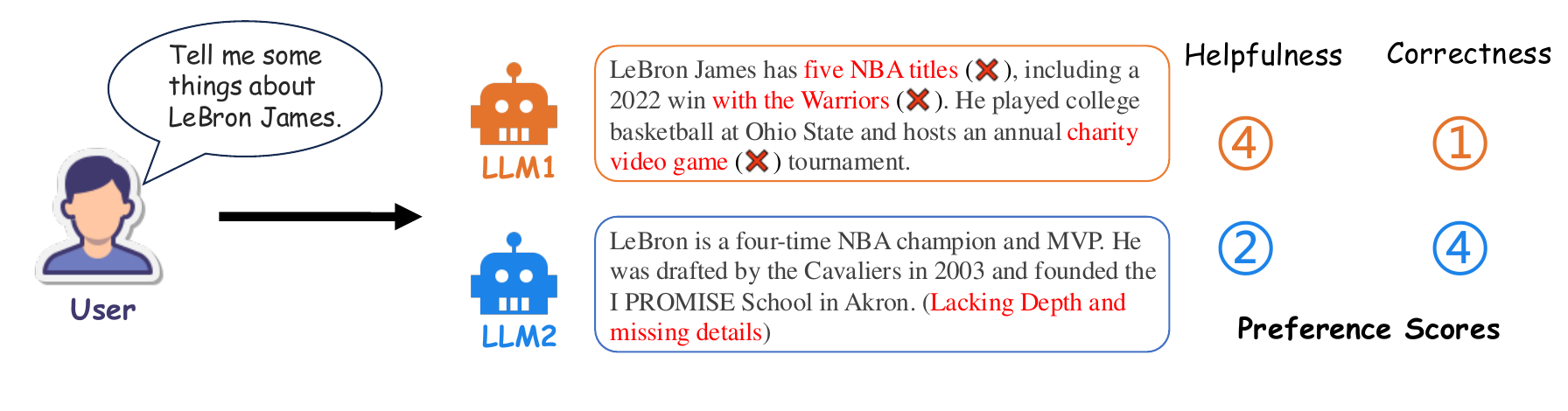}
    \caption{\hl{LLM1 prefers helpfulness while LLM2 prefers correctness, yet both struggle to adapt across various dimensions.}}
    \label{fig:intro}
\end{figure}

\hl{Recent breakthroughs in Large Language Models (LLMs)}~\citep{gpt42023openai,llama3,qwen2.5} \hl{have intensified the need for alignment with multi-dimensional human preferences, which is a critical requirement for deploying safe and controllable AI systems.}
\hl{While preference optimization methods like RLHF}~\citep{ouyang2022training} \hl{and DPO series}~\citep{DPO2023Rafailov,ethayarajh2024kto,ipo2024Azar,CPO2024xu} \hl{have advanced single-objective alignment, fundamental limitations persist when addressing the multi-dimensional nature of real-world human preferences}~\citep{helpsteer22024wang,ultrafeedback2024cui,aaai2025xuderong}.
\hl{Figure}~\ref{fig:intro} \hl{shows that different LLMs favor different criteria but struggle to balance these dimensions, highlighting the need for more comprehensive multi-objective alignment methods.}

\hl{Existing multi-objective alignment methods} can be categorized into two main branches.
On the one hand, instruction-controlled methods~\citep{ControlPO2024guo,JANUS2024Lee,RiC2024yang,MCA2024fu}~insert control tokens into the prompt for each preference. 
However, these methods are constrained by predefined prompts and lack adaptability in aligning with various dimensions.
On the other hand, multi-model integration methods involve training multiple reward models, or even language models, for each preference dimension. For example, methods like MORLHF~\citep{zhou2023morlhf} and Reward Soups~\citep{rame2024rewarded} require training multiple reward models, while MODPO~\citep{MODPO2024zhou} integrates multiple preference objectives into the margin calculation of the reward model. HM3~\citep{zhou2024hm3} uses model merging to combine pre-trained models across different dimensions. Obviously, these methods are time-consuming and costly.

To alleviate these issues, we aim to reduce the reliance on multiple models and enhance adaptability to multiple dimensions
by proposing Adaptive Multi-objective Preference Optimization (\ourtitle).
Firstly, we propose a multi-objective optimization framework by designing the multi-dimensional alignment model, enabling it to focus more precisely on each dimension. By integrating generation metrics \citep{simpo2024meng} of language models as implicit rewards, our framework allows LLMs to inherently understand and align with the meaning associated with different dimensions.
Furthermore, we develop an adaptive weight assignment mechanism that automatically determines the importance of each dimension based on the modeled output space. The contributions of this paper can be summarized as follows:
\begin{itemize}[itemsep=0pt, topsep=0pt]
    \item \hl{We propose a multi-objective optimization framework using the Multi-objective BT model, eliminating the need for additional reward or reference models. This approach integrates a multi-dimensional paradigm with generation metrics as implicit rewards to balance various dimensions, ensuring a lightweight and efficient framework.}
    \item \hl{Building on the aforementioned framework, we introduce an adaptive weight assignment mechanism that models the generation space of LLMs as a Gaussian distribution. Sampling from this distribution dynamically determines the importance weight for each dimension, allowing~\ourtitle~to adaptively perceive significance levels across dimensions.}
    \item Empirical evaluations against multiple state-of-the-art (SOTA) baselines across several benchmarks demonstrate the effectiveness and superior performance of our proposed~\ourtitle. \hl{Further experiments on different scales and series of models confirm the effectiveness and scaling ability of~\ourtitle.}
\end{itemize}

%% file: latex/sections_article/2_related_rjq.tex

\subsection{Preference Alignment}
Research on preference alignment can be categorized into two main classes. One line of research focuses on Reinforcement Learning with Human Feedback (RLHF). \citet{ouyang2022training}~demonstrated the effectiveness of RLHF for instruction-following models. Extensions~\cite{touvron2023llama2,yuan2023rrhf,wen2024language,li2024safety}~have emerged as the leading strategy to integrate human preferences via RLHF.
However, RLHF faces challenges due to the instability of Proximal Policy Optimization (PPO) and the sensitivity of reward model training.
To address these issues, another line of research has been proposed that eliminates the need for reinforcement learning by directly optimizing preferences~\cite{DPO2023Rafailov,KTO2024Ethayarajh,rdpo2024Park,ipo2024Azar,caldpo2024xiao}. Nonetheless, these methods often involve multiple models, leading to high computational costs.

\subsection{Reference-Free Preference Alignment}
Variants of the DPO series require additional memory to deploy the reference model during training. 
To address this issue, CPO~\citep{CPO2024xu}~demonstrates that the loss of DPO can be effectively approximated using a unified reference model. SimPO~\citep{simpo2024meng}~replaces the reward in DPO with a sequence generated by the policy model that approximates the maximization of the mean log-likelihood, serving as the implicit reward. ORPO~\citep{ORPO2024Hong}~explores another formulation of the implicit reward.
These methods improve the computational and memory efficiency of LLM training. However, they are primarily focused on single-dimensional preference optimization and do not sufficiently address the intrinsically diverse nature of human preferences.

\subsection{Multi-objective Preference Alignment}

Multi-objective alignment methods have been developed in recent years to accommodate diverse user preferences. 
A series of studies have extended RLHF and DPO into multi-objective frameworks, such as MORLHF~\citep{zhou2023morlhf}~and MODPO~\citep{MODPO2024zhou}. This involves multiple models, including reward models, policy models, or reference models. Some extensions~\citep{emmerich2018tutorial, mukherjee2024multi,rewardsoup2023Rame,zhou2024hm3,RLPHF_2023jang,LIMA2023zhou,DPA2024wang,HPO2024Badrinath} also aim to adjust a series of specialized models for each alignment dimension, but these approaches incur significant computational costs. 
On the other hand, another series of work~\citep{ControlPO2024guo,JANUS2024Lee,RiC2024yang,MCA2024fu}~focuses on an instruction-controlled approach. These methods incorporate user preferences as a "control token`` within the prompt, achieving controllability and reducing the number of trained models to just one. However, these approaches often lack flexibility and adaptability in aligning with various dimensions. 
In contrast to the above methods, we employ only a single policy LLM to achieve adaptive multi-objective optimization.

%% file: latex/sections_article/3_pre_rjq.tex



\paragraph{Bradley-Terry (BT) Reward Model.}


Given an LLM $\pi_{\theta}$ parameterized by $\theta$, query $x$, and response $y$, classical preference alignment methods can be formulated as:
\begin{equation}
\label{eq.1}
\small
\max_{\pi} \mathbb{E}_{\substack{x \sim \mathcal{D} \\ y \sim \pi(\cdot \mid x)}} \left[ r(y \mid x) \right] - \beta D_{\text{KL}} \left[ \pi_{\theta}(y \mid x) \| \pi_{\text{ref}}(y \mid x) \right],
\end{equation}
where $r$ measures the consistency with human preference, $\mathcal{D} = \{(x,y_w,y_l)^i\}$ denotes the preference dataset that prefers $y_w$ and disprefer $y_l$, $D_{\text{KL}} \left[ \pi(y \mid x) \| \pi_{\text{ref}}(y \mid x) \right]$ is the Kullback-Leibler divergence~\cite{rl2007jan,position2024Reizinger} between $\pi$ and reference LLM $\pi_{ref}$, and $\beta$ is a constant controlling the strength of KL.

In measuring human preferences, one of the most commonly used approximations is the Bradley-Terry (BT) reward model~\cite{bradley1952rank,sun2024rethinking,guo2024todo}. Assumed $r^*$ is the latent ground-truth reward function, the BT model defines the preference distribution $\pi^*$ as:
\begin{equation}
\small
    \label{eq.2}
    \pi^*(y_w \succ y_l \mid x) = \frac{\exp(r^{*}(x, y_{w}))}{\exp(r^{*}(x, y_{w})) + \exp(r^{*}(x, y_{l}))}.
\end{equation}

\paragraph{Simple Preference Optimization (SimPO).}
Compared with the primary methods~\cite{DPO2023Rafailov}, SimPO~\cite{simpo2024meng} introduces a more lightweight and straightforward approach that eliminates the need for a reference model. SimPO utilizes the generation metric~\cite{gpt42023openai,radford2018improving,touvron2023llama,touvron2023llama2,llama3}, i.e, the average log-likelihood of generating the output sequence $y$, which consists of $m$ tokens given an input sequence $x$, is computed as:
\begin{equation}
\small
    \label{eq.3}
    \log \pi_{\theta}(y \mid x) = \frac{1}{m} \sum_{t=1}^{m} \log \pi_{\theta}(y_{t} \mid x, y_{<t}).
\end{equation}

Thus, replacing the reward with Eq.\eqref{eq.3},  the optimization objective of SimPO can be formulated as:
\vspace{-0.2em}
\begin{equation} 
\small
\label{Eq.4} 
\begin{split} 
\mathcal{L}_{\text{SimPO}}(\pi_\theta) = -\mathbb{E}_{(x, y_w, y_l) \sim \mathcal{D}} \biggl[ \log \sigma \biggl( \frac{\beta}{|y_w|} \log \\ 
\pi_\theta(y_w \mid x) - \frac{\beta}{|y_l|} \log \pi_\theta(y_l \mid x) - \gamma \biggr) \biggr], \end{split} 
\end{equation}
where $\beta$ is a constant that controls the scaling of the reward difference and $\gamma$ is a target reward margin.

\paragraph{Multi-objective Preference Dataset.}
A $K$-dimensional preference dataset~\cite{ultrafeedback2024cui,helpsteer2024wang,helpsteer22024wang} can be divided into subsets \(\mathcal{D}_j\) based on objective \(j\), represented as \(\bm{\mathcal{D}} = [ \mathcal{D}_1, \mathcal{D}_2, \ldots, \mathcal{D}_K ]\). Here, each subset including $N$ samples, denoted as \(\mathcal{D}_j = \{ (x, y_w, y_l, d_j)^i \}_{i=1}^{N}\), where \({d_j}^{i}\) represents the score of the \(i\)-th sample in the dimension \(j\), such as helpfulness. 
This structure facilitates the assessment of preferences across multiple dimensions by organizing data according to specific objectives.

\paragraph{Multi-objective Preference Optimization (MOPO).}
Classical MOPO methods~\cite{2DDPO2024li,MODPO2024zhou} use separate reward models $\bm{r}^*={[{r_1}^*,...,{r_K}^*]}$ for multi-dimensional datasets $\bm{\mathcal{D}} = [ \mathcal{D}_1, \mathcal{D}_2,...,\mathcal{D}_K ]$, allowing customization of LLMs by adjusting reward weightings.
In the context of multi-objective alignment, the primary goal is to approximate a Pareto front~\cite{mukherjee2024multi,safest2024sinha} that represents the multi-dimensional preferences for LLMs. Each preference is optimized according to a specific collective reward model $r^*_i$, with an associated known weight \(\alpha_i\).
The overall objective for MOPO is to maximize the cumulative weighted sum of rewards across all dimensions. This can be expressed as:
\begin{equation}
\small
    \label{eq.5}
    \max_{\pi} \mathbb{E}_{\substack{x \sim \bm{\mathcal{D}} \\ y \sim \pi(\cdot \mid x)}} \left[ R(y \mid x, d_1,...,d_K) \right] ,
\end{equation}
where \(R\) represents the dot product of the vector \(\bm{\alpha} = [\alpha_1, ..., \alpha_K]\) and $\bm{r}^*={[{r_1}^*,...,{r_K}^*]}$.

Generally, in the context of multi-objective alignment, the goal is to approximate a Pareto front of the multi-dimensional preference for LLMs.
Each preference is optimized for a specific collective reward model $r^*_i$, with an associated known weight \(\alpha_i\).
The weighted sum of these auxiliary rewards is expressed as \(R(x, y)\), represented by the dot product of the vector \(\bm{\alpha} = [\alpha_1, ..., \alpha_K]\) and $\bm{r}^*={[{r_1}^*,...,{r_K}^*]}$, denoted as follows.
\begin{equation}
\label{eq.6}
\small
    R(y \mid x, d_1,...,d_K) = \sum_{j=1}^{K} \alpha_{j} r^*_{j}(x, y, d_j) = \bm{\alpha}^{\top} \mathbf{r}_{xy}.
\end{equation}

%% file: latex/sections_article/4_RMoPO_rjq.tex





\begin{figure*}[!h]
	\centering
	\includegraphics[width=0.93\textwidth]{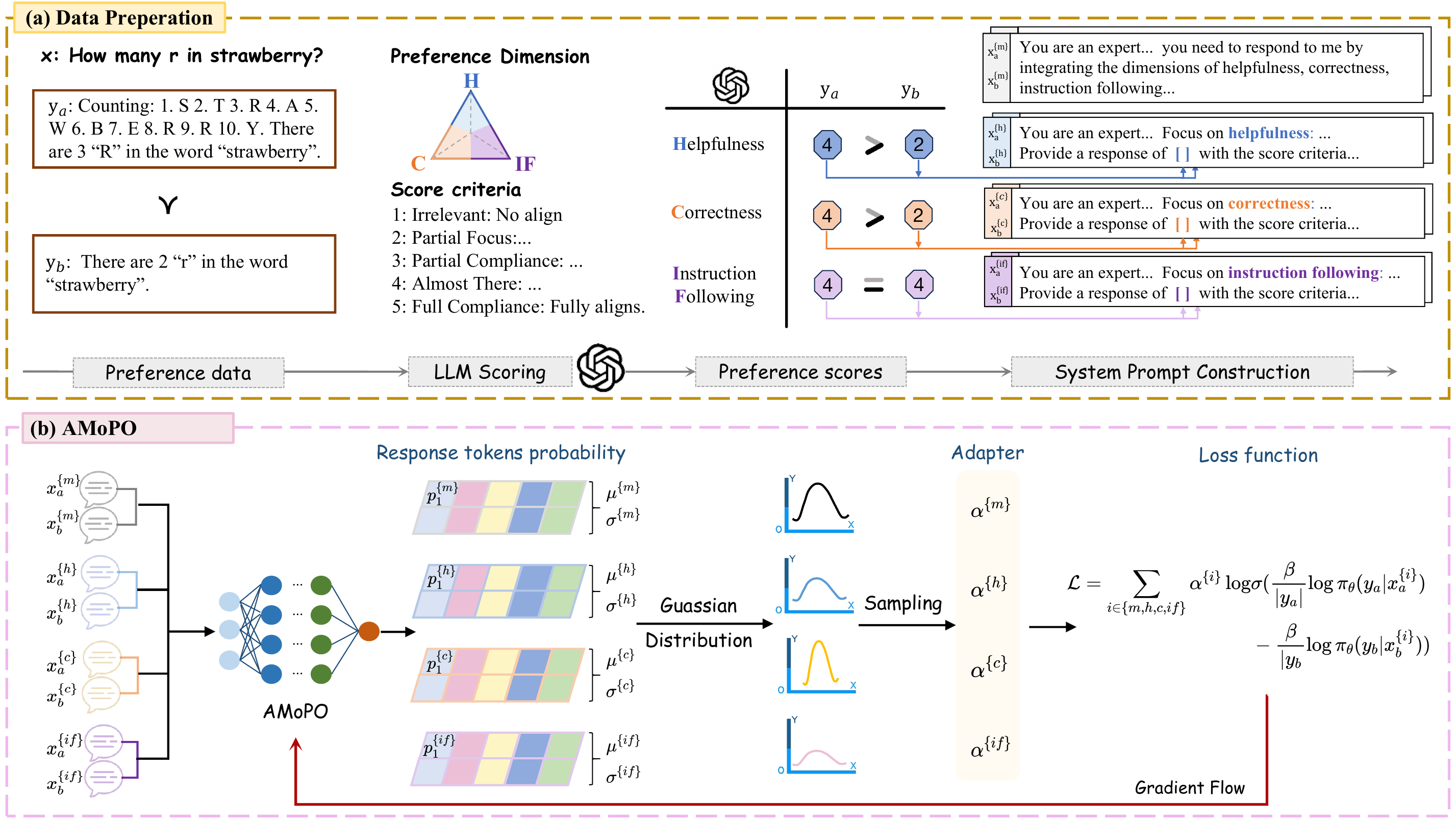}
    \caption{The pipeline of \ourtitle~comprises two key stages: (a) \textbf{Data Preparation}: For a prompt with two responses, $y_a$ and $y_b$, GPT-4 scores each response using the scoring criteria from three dimensions: helpfulness, correctness, and instruction following. For each dimension, a unique template is designed to ensure the model focuses on that specific dimension when generating the answer. (b) \textbf{Training}: A Gaussian-based weight assignment mechanism is employed to sample weights for each dimension. \ourtitle~is then updated using the loss $\bm{\mathcal{L}}$. }
    \label{fig:pipeline}
\end{figure*}

This section begins with the process of data preparation. We then present our~\ourtitle~by exploring the multi-objective derivation and the adaptive weight assignment mechanism. The complete pipeline of our method is illustrated in Figure~\ref{fig:pipeline}.

\subsection{Data Preparation}
\label{sec:data_pre}
Figure~\ref{fig:pipeline}~(a) illustrates that for a given prompt with two responses, $y_a$ and $y_b$, GPT-4 scores each response along $K$ dimensions: helpfulness, correctness, and adherence to instructions. For the detailed scoring template, please see Appendix~\ref{prompts:3dimensions}.

By leveraging these scores $\bm{d}=[d_1....,d_K]$, we can develop a novel prompt template that integrates both the dimension and the corresponding score, a process denoted as $x^* = f(x, d)$.
This approach guides the model to focus on the specified dimension when generating a response. A preference dataset with $K$ dimensions can be partitioned into subsets \(\mathcal{D}_j\) according to objective \(j\), expressed as \(\bm{\mathcal{D}} = [ \mathcal{D}_1, \mathcal{D}_2, \ldots, \mathcal{D}_K ]\). Each subset comprises $N$ samples, represented as \(\mathcal{D}_j = \{ (x, y_a, y_b, d_j)^i \}_{i=1}^{N}\), where \({d_j}^{i}\) indicates the score of the \(i\)-th sample in dimension \(j\), such as helpfulness. 


\subsection{Multi-objective Optimization Derivation}

\input{latex/images/main/amopo_algorithm}

\paragraph{Overall Optimization Objective.}
\hl{Here, we propose a novel paradigm based on weighted sum multi-objective optimization~\citep{miettinen1999nonlinear}, which constructs a scalar objective function to find the optimal solution. Specifically, we design a weight vector $\bm{\alpha} = [\alpha_1, \ldots, \alpha_K]$, obtained through the policy} $\alpha \sim \rho$, where $\sum_{k=1}^{K} \alpha_k = 1$ and $\alpha_k > 0$.
The Pareto solution is then determined by addressing the following scalar optimization problem:
\begin{equation}
    \normalsize
    \label{eq.overall}
    \begin{split}
        & \text{max} \  \pi(y_w \succ y_l | x, \bm{\alpha}, \bm{d}) \\
        & \rightarrow \text{max}  \  {\textstyle \sum\limits_{k=1}^{K} } \rho(\alpha_k|x_k^*) \log \pi^*_k(y_w \succ y_l | x_k^*),
    \end{split}
\end{equation}
\hl{where we restructure} ${x}^*_k = f(x, {d_k})$ to adapt to the $k^{th}$ dimension in Section~\ref{sec:data_pre}, and $\pi^*_k$ denotes the preference alignment function for the $k^{th}$ dimension.



\hl{To optimize and balance multiple objectives effectively, dynamic adjustments are essential. Fixed weights often miss the complexities and trade-offs of different dimensions. Dynamic weight sampling allows for adaptive prioritization based on current conditions, ensuring the process remains efficient and relevant. Thus, optimizing both the weight policy $\bm{\alpha} \sim \rho$ and the preference alignment policy $p^*$ is crucial, enabling the framework to maintain balance among objectives in varying scenarios.}

\paragraph{Multi-Objective BT (MOBT) Model.}
\hl{
We introduce the Multi-Objective BT (MOBT) model to optimize the preference alignment policy, marking a significant departure from the traditional BT model described in Eq.\eqref{eq.2}. Unlike the conventional approach, which fuses scores from multiple dimensions into a singular $r(x,y)$, our model employs a dynamically weighted sum of scores across various dimensions. 
This innovative strategy facilitates adaptive learning of importance weights, thereby balancing scores across multiple dimensions and achieving an adaptive multi-objective optimization framework.} This can be formulated as follows:
\begin{equation}
    \normalsize
    \label{eq.7}
    \begin{split}
    & \pi(y_w \succ y_l \,|\, x;\, \bm{\alpha};\, \bm{d} ) \\
    & = \sum_{k = 1}^{K} \alpha_{k} \log \sigma \left(  r_{k}(x^{*}_k, y_w) - r_{k}(x^{*}_k, y_l) \right) \\
    & =  \log \prod_{k=1}^{K}  \sigma^{\alpha_k}(r_{k}(x^{*}_k, y_w) - r_{k}(x^{*}_k, y_l)) \\
    & \text{s.t.} \sum_{k=1}^{K} \alpha_k = 1,~x^{*}_k = f(x, {d_k}).
    \end{split}
\end{equation}
In our model, $\alpha$ is derived from the weight policy $ \rho$, and $\sigma(x) = \frac{1}{1 + \exp(-x)}$ is the Sigmoid function. 
Evidently, the multi-objective BT Model reduces to the traditional BT Model when $K = 1, \alpha=1$.

\paragraph{{\ourtitle} Loss.}
Let the LLM policy be $\pi_\theta$ and the adaptive weight assignment policy be $\rho_{\varphi}$, we can formulate a maximum likelihood estimation objective by combining Eq.\eqref{eq.3}, Eq.\eqref{eq.7} and Eq.\eqref{eq.10} on the multi-dimensional dataset $\mathcal{D}$ as follows.
\begin{equation}
\normalsize
\label{Eq.ours}
    \begin{split}
        \mathcal{L}&_{\text{\ourtitle}}(\pi_\theta; \bm{\alpha}; \bm{d}) = 
        - \mathbb{E}_{(x, y_w, y_l)\sim \mathcal{D}} \biggl[ \sum_{k=1}^{K} \alpha_k \log \sigma \\ & \biggl(\frac{\beta}{|{y_w}|} \log \pi_\theta({y_w} \mid x^{*}_k) \quad - \frac{\beta}{|{y_l}|} \log \pi_\theta({y_l} \mid x^{*}_k) \biggr) \biggr]\\
        &~~~~~~~~~~~ \text{s.t.}\; x^{*}_k=f(x,{d_k}) ,
    \end{split}
\end{equation}
The derivations are detailed in Appendix~\ref{math_derivation}.

\subsection{Adaptive Weight Assignment Mechanism}
To dynamically balance the contributions of each objective dimension, we utilize the language model's token generation probabilities to compute the mean $ \mu_k $ and variance $ \sigma_k^2 $ for each dimension $ k $.
These are used to parameterize a Gaussian distribution $ \rho_k $, from which the weight $ \alpha_k $ is sampled. This allows the model to adjust each dimension's significance based on confidence in token generation~\cite{shannon1948mathematical, Maximum1983Bahl}.
Formally, the mean and variance are:
\begin{equation}
\normalsize
    \label{eq.10}
    \mu_k = \frac{1}{T} \sum_{t=1}^{T} p_{k,t}, \ \sigma_k^2 = \frac{1}{T} \sum_{t=1}^{T} (p_{k,t} - \mu_k)^2,
\end{equation}
where $ p_{k,t} = \pi_{\theta}(y_{t} \mid x_k^*, y_{<t}) $ is the probability of the $ t^{th} $ token for dimension $ k $.
Then the weight $ \alpha_k $ is sampled as: $\alpha_k \sim \mathcal{N}(\mu_k, \sigma_k^2)$.
We normalize all the weights using the formula \(\alpha_i = \frac{e^{\alpha_i}}{\sum_{j=1}^{n} e^{\alpha_j}}\) to ensure that \(\alpha_k > 0\) and \(\sum_{k=1}^{K} \alpha_k = 1\). 

This adaptive mechanism enhances the model’s ability to adaptively prioritize dimensions with higher confidence.
Notably, the process of generating $\alpha$ is modular and interchangeable, allowing for any desired refinements or mechanisms. This flexibility ensures that the weight vector $\bm{\alpha}$ dynamically adjusts based on model performance across dimensions, facilitating balanced optimization. The versatility of this mechanism will be further detailed in the experimental section.

\subsection{Implementation details}

As shown in Algorithm~\ref{alg:amopo},
{\ourtitle} transfers an original quadruple $<x,y_w,y_l,\bm{d}>$ into a set of quadruples $\{<x,y_w,y_l,d>\}^{d \in \bm{d}}$. And then we design a map function $f$, which changes $x$ according to different dimensions, i.e., $x^* = f(x,d)$. Subsequently, we concatenate the multi-dimensional prompts $x^*$ with the winning response $y_w$ and the losing response $y_l$ as inputs to the policy model $\pi_\theta$, obtaining the $p_d^w$ and $p_d^l$ individually.
We model the sum probabilities $p_d$ through Gaussian Sampling, deriving the weight $\alpha_d$.
By combining the weights of each dimension with the generated metrics, we can obtain the final expression for {\ourtitle}.


%% file: latex/images/main/amopo_algorithm.tex
\begin{algorithm*}[!htbp]

\caption{AMoPO}
\label{alg:amopo}
\setstretch{0.2}
\small{
\KwIn{Policy model $\pi_\theta$, Objectives $\bm{d}$, Mapping $f$, Epochs $epochs$, Coefficient $\beta$, Learning rate $\eta$, Dataset $\mathcal{D} = \{ (x, {y_w}, {y_l}, \bm{d})^i\}_{i=1}^{N}$}
\For{epoch \textnormal{in} range($epochs$)}{
    {\textbf{Initialize:} All tokens probabilities $\bm{p}= [\;]$, Weights $\bm\alpha=[\;]$, Multiple loss $\bm{\mathcal{L}} = [\;]$}\\
    Sample a batch $\mathcal{D}_s = \{ (x, {y_w}, {y_l}, \bm{d})^s\}_{s=1}^{S}$ from $\mathcal{D}$;\\
    
    \For{$d$ \textnormal{in} $\bm{d}$}{
        Use the dimension $d$ and the mapping function $f$ to get the unique prompt template $x^* = f(x,d)$;\\
        Merge all samples $\mathcal{D}_j^d = \{ (x^*, {y_w},{y_l}, d_{score})^s\}_{s=1}^{S} $; \\
        Predict the all tokens probabilities $p_d^{w}={\{<t_i^w,p_i^w>\}}_{i=1}^{|y_w|}$ and $p_d^{l}={\{<t_i^l,p_i^l>\}}_{i=1}^{|y_l|}$ based on $\pi_\theta(y_w|x^*)$ and $\pi_\theta(y_l|x^*)$;\\
        Calculate the sum probabilities $p_d = p_d^w+p_d^l$;\\
        Model $p_d$ based on Gaussian Sample to get the weight $\alpha_d$;\\
        Calculate current function $l=\log\sigma(\frac{\beta}{|{y_w}|} \log \pi_\theta({y_w} \mid x^{*})-\frac{\beta}{|{y_l}|} \log \pi_\theta({y_l} \mid x^{*}))$;\\
        $\bm{p} \gets p_d$ ;
        $\bm\alpha \gets \alpha_d$;
        $\bm{\mathcal{L}} \gets l$;\\
    }
    Normalize different dimensions weights $\bm{\alpha}$ ;\\
    $\theta \gets \theta+\eta{\nabla_\theta}\mathbb{E}_{(x, y_w, y_l,\bm{d})\sim \mathcal{D}_s} \bm{\alpha}^{T}\bm{\mathcal{L}}  $
}
\KwOut{$\pi_\theta$}
}
\end{algorithm*}

%% file: latex/sections_article/5_Experiment.tex

\subsection{Experiment Setup}

\paragraph{Training Setup.}We perform preference optimization based on HelpSteer2~\cite{helpsteer22024wang}.
It not only includes instructions and preferred responses, but it also scores each response across multiple dimensions such as helpfulness, correctness and etc. While some researches~\cite{complexbench2024wen, ORPO2024Hong} have confirmed that instruction following is an indispensable dimension for evaluating the quality of a response, we leverage \textit{gpt4o-2024-05-13} to evaluate the instruction-following scores for each response. The template is shown in Appendix ~\ref{prompts:instruction_following}.

\paragraph{Models.}We perform preference optimization with two models from Qwen family~\cite{qwen2.5}, Qwen2.5-7B-Instruct and Qwen2.5-14B-Instruct. Both models have demonstrated robust capabilities~\cite{qwen2, qwen2.5, qwen2.5code2024hui, wu2024beyond} across various tasks, showing strong performance whether fine-tuned or used directly. 
We evaluate the effectiveness of our preference optimization method on these two models with different levels of comprehension.



\begin{table}[!htbp]
\centering
  \caption{Evaluation details on benchmarks.}
  \label{tab:evaluation_details}

\resizebox{0.48\textwidth}{!}{
\begin{tabular}{ccccc}
\toprule

{\LARGE \textbf{Dataset}}& \LARGE Examples  & \LARGE baseline model &\LARGE Judge Model & \LARGE Metric \\
\midrule
 \LARGE AlpacaEval2.0 & \LARGE 805 & \LARGE GPT-4\footnotemark[1]  & \LARGE GPT-4o\footnotemark[3] & \LARGE Win Rate \\
 \LARGE ArenaHard & \LARGE 500 &\LARGE GPT-4\footnotemark[2] &\LARGE GPT-4o\footnotemark[3]& \LARGE Win Rate \\
 \LARGE MT-bench&\LARGE 80 & \LARGE - &\LARGE GPT-4o\footnotemark[3] &\LARGE Rating(1-10)\\
\bottomrule
\end{tabular}
}
\end{table}

\paragraph{Benchmarks.}We mainly test our method on three of the most popular and recognized instruction-following benchmarks: AlpacaEval2.0~\cite{alpaca_eval}, Arena-Hard~\cite{arenahard2024} and MT-bench~\cite{MT-bench2024zheng}. These benchmarks consist of a set of queries and their answers are under the framework of LLM-as-a-Judge~\cite{MT-bench2024zheng} (Details are in Table \ref{tab:evaluation_details}).

\paragraph{Baselines.}We divide the compared methods into two categories: single-objective and multi-objective preference optimization.
For the \textbf{{single-objective preference optimization}} methods:
(1) DPO~\cite{DPO2023Rafailov} eliminates the training of additional reward functions, which directly learn a policy model from preference data.
(2) Cal-DPO~\cite{caldpo2024xiao} calibrates the implicit reward function against the ground-truth rewards based on DPO.
(3) ORPO~\cite{ORPO2024Hong} introduces a reference-free model odd ratio to compare winning and losing responses and jointly trains with the SFT stage.
(4) SimPO~\cite{simpo2024meng} proposes to leverage the generation metric with the policy model to present the implicit reward for responses, significantly simplifying the training phase.
For the \textbf{{multi-objective preference optimization}} methods: 
(1) MODPO~\cite{MODPO2024zhou} divides the multi-objective training into two steps: For two dimensions, e.g., helpfulness and correctness, MODPO firstly trains a reward model in correctness dimension. Then, it introduces scores by the trained reward model when training on the helpfulness dimension.
(2) CPO~\cite{ControlPO2024guo} \hl{extends both SFT and DPO to solve the multi-objective optimization problem.}
(3) SPO~\cite{SPO2024Lou}
\hl{sequentially fine-tunes LLMs to align with multiple dimensions of human preferences.}
Appendix~\ref{imple_details} shows more details.

\footnotetext[1]{Version: GPT-4-preview-1106.}
\footnotetext[2]{Version: GPT-4-0314.}
\footnotetext[3]{Version: GPT-4o-2024-05-13.}
\subsection{Main Results}
As shown in Table~\ref{tab:performance_comparison}, our approach achieves 28.5$\%$ overall improvements over the baseline methods, benefiting from our multi-dimensional alignment mechanism. 
\begin{table*}[h]
\centering
\resizebox{0.9\textwidth}{!}{
\begin{tabular}{llcccccccc}
\toprule
& & \multicolumn{3}{c}{Qwen2.5-7B-Instruct}  & \multicolumn{3}{c}{Qwen2.5-14B-Instruct} \\
\cmidrule(lr){3-5} \cmidrule(lr){6-8}
\textbf{Objective} & \textbf{Method} & AlpacaEval 2 & Arena-Hard & MT-Bench & AlpacaEval 2 & Arena-Hard & MT-Bench  \\
 \cmidrule(lr){3-3} \cmidrule(lr){4-4}
\cmidrule(lr){5-5} \cmidrule(lr){6-6} \cmidrule(lr){7-7} \cmidrule(lr){8-8}
& & WR (\%) & WR (\%) & Avg.Rating & WR (\%) & WR (\%) & Avg.Rating& \\
\midrule
\multirow{4}{*}{\textbf{Single}} &
DPO &41.5 & 56.4 & 8.58 & 56.4 & 66.8 & 8.77 \\
& ORPO & 13.1 & 27.4 & 7.92 & 18.8 & 34.2 & 8.07 \\
& SimPO & \underline{56.1} & \underline{64.2} & 8.50 & 58.6 & 67.8 & \underline{8.90} \\
& Cal-DPO & 41.7 & 55.4 & \underline{8.60} & 52.8 & 65.8 & 8.68 \\
\midrule
\multirow{4}{*}{\textbf{Multiple}} &
\hl{CPO}  & \hl{52.0} & \hl{62.0} & \hl{8.43} & \underline{\hl{59.0}} & {\hl{69.4}} &  \hl{8.8} \\
& SPO & 41.2 & 60.4 & 8.58 & 55.9 & \underline{71.2} & 8.8 \\
& MODPO & 52.8 & 63.4 & 8.57 & 55.7 & 68.8 & 8.72 \\
& \ourtitle(ours) & \textbf{58.5} & \textbf{64.4} & \textbf{8.63} & \textbf{63.9} & \textbf{73.4} & \textbf{9.0} \\
\bottomrule
\end{tabular}
}
\caption{Performance comparison of different methods on Qwen2.5-7B-Instruct and Qwen2.5-14B-Instruct across various benchmarks.}
\label{tab:performance_comparison}
\end{table*}
For \textbf{Qwen2.5-14B-Instruct} as the base model, 
{\ourtitle} surpasses the current state-of-the-art methods on three datasets. \hl{It outperforms CPO by up to 4.9 points on AlpacaEval 2 and outperforms SPO by up to 2.2 points on ArenaHard.} And it also outperforms SimPO by up to 0.1 average rating improvement on MT-bench.
For single-objective preference optimization methods, e.g., SimPO, these methods do not generate responses from multidimensional perspectives.
\hl{For current multidimensional preference optimization methods, e.g., CPO and SPO, these methods do not incorporate the meanings of the dimensions into the model training.}
To tackle the above issues, {\ourtitle} integrates the significance of multiple dimensions into the preference optimization and gives alignment templates to activate the model's learning.
In addition, compared with MODPO, {\ourtitle} does not require any additional training stage to alleviate extra computation overhead.


For \textbf{Qwen2.5-7B-Instruct} as the base model,  
{\ourtitle} still outperforms SimPO by up to 2.4 points on AlpacaEval 2, 0.2 points on Arena-Hard, and outperforms Cal-DPO by up to 0.03 average rating on MT-bench. 
We observe that the improvement compared to the results of the 14B model is relatively constrained. 
The reason is that the key design of {\ourtitle} is to make the policy model align different dimensions with score criteria during preference optimization. 
With fewer policy model parameters, the understanding and studying capabilities are greatly weakened.
We find that the response of the trained model based on Qwen2.5-7B-Instruct will return unexpected information or hallucinated details compared to Qwen2.5-14B-Instruct.
The details are shown in Appendix~\ref{Cases}.
\hl{In addition, we also recognize that the performance is not strong as expected on ArenaHard. 
This can be explained by the already impressive performance of the Qwen2.5-7B-Instruct model, which leaves little room for further enhancement on more demanding benchmarks like ArenaHard.
To further validate the efficiency of ~\ourtitle, we extend LLaMA3.1-8B-Instruct with slightly weaker capabilities than Qwen2.5-7B-Instruct but similar parameter sizes.
AMoPO has been proven to deliver substantial gains even when applied to LLaMA3.1-8B-Instruct.}
The detailed analysis and results are shown in Appendix.~\ref{differenrt_model}.

To more intuitively observe the performance differences, we also report the win rate between {\ourtitle} and other methods. 
\hl{Experimental results demonstrate that the {\ourtitle} has better response quality, and achieves 81.5$\%$ average win rate} (Details are in Appendix.~\ref{appendix:more_results}).

\subsection{Different Weighted Policies}

\hl{To delve into the effectiveness and generalization performance of ~\ourtitle, we design three groups of experiments on three benchmarks.}
\hl{Table}~\ref{tab:performance_comparison_weighted} \hl{presents the results. 
The Gaussian Sample strategy achieves 1.73 points, 1.13 points, and 0.14 rating average improvement than fixed weight with concise response, verifying the efficiency and superiority.
The decay of the challenging ArenaHard on the 7B model indicates that the lower ability model might be unable to determine the most appropriate course of action during evolution.
As the number of parameters increases, e.g., Qwen2.5-32B-Instruct, our approach achieves greater gains compared to fixed weight without generating redundant responses, demonstrating that the model possesses the ability to identify the most suitable direction for growth.
With the increase in model size, the design of {~\ourtitle} encourages the model itself to explore more space to adaptively learn the multidimensional aspects, enhancing the capabilities.}

\begin{table*}[!t]

\centering
    
\resizebox{0.9\textwidth}{!}{
\begin{tabular}{llcccccc}
\toprule
\multirow{2}{*}{\textbf{Model}} & \multirow{2}{*}{Weighted Policy} & \multicolumn{2}{c}{AlpacaEval 2} & \multicolumn{2}{c}{Arena-Hard} & \multicolumn{2}{c}{MT-Bench} \\
\cmidrule(lr){3-4} \cmidrule(lr){5-6} \cmidrule(lr){7-8}
& & WR (\%) & Avg.Len  & WR (\%) & Avg.Len & Rating & Avg.Len \\
\midrule
\multirow{2}{*}{\hl{Qwen2.5-7B-Instruct}} 
& \hl{+Fixed Weight(1:1:1:1)} & \hl{57.6(+0.9)} & \hl{1602} & \hl{\textbf{65.2}(-0.8)} & \hl{1994} & \hl{8.44(+0.19)} & \hl{\textbf{1408}} \\
& \hl{+Gaussian Sample} & \hl{\textbf{58.5}} & \hl{\textbf{1600}} & \hl{64.4} & \hl{\textbf{1989}} & \hl{\textbf{8.63}} & \hl{1415} \\

\cmidrule{1-8}
\multirow{2}{*}{\hl{Qwen2.5-14B-Instruct}} 
& \hl{+Fixed Weight(1:1:1:1)} & \hl{63.6(+0.3)} & \hl{1521} & \hl{72.8(+0.6)} & \hl{2014} & \hl{8.97(+0.03)} & \hl{1342} \\
& \hl{+Gaussian Sample} & \hl{\textbf{63.9}} & \hl{\textbf{1518}} & \hl{\textbf{73.4}} & \hl{\textbf{2013}} & \hl{\textbf{9.0}} & \hl{\textbf{1327}} \\

\cmidrule{1-8}
\multirow{2}{*}{\hl{Qwen2.5-32B-Instruct}}
& \hl{+Fixed Weight(1:1:1:1)} & \hl{65.3(+4.0)} & \hl{1649} & \hl{88.4(+3.6)} & \hl{2177} & \hl{9.1(+0.2)} & \hl{1585} \\
& \hl{+Gaussian Sample} & \hl{\textbf{69.3}} & \hl{\textbf{1634}} & \hl{\textbf{92.0}} & \hl{\textbf{2166}} & \hl{\textbf{9.3}} & \hl{\textbf{1535}} \\
\bottomrule
\end{tabular}
}
\caption{\hl{Performance comparison of different weighted policies within the MOBT framework on Qwen2.5-7B/14B/32B-Instruct models}}
\label{tab:performance_comparison_weighted}
\end{table*}

\subsection{Detailed Analysis}
We analyze relationships among dimensions by tracking helpfulness, correctness, and instruction following margins during training. 
\begin{figure}[h]
    \centering
    \includegraphics[width=0.36\textwidth]{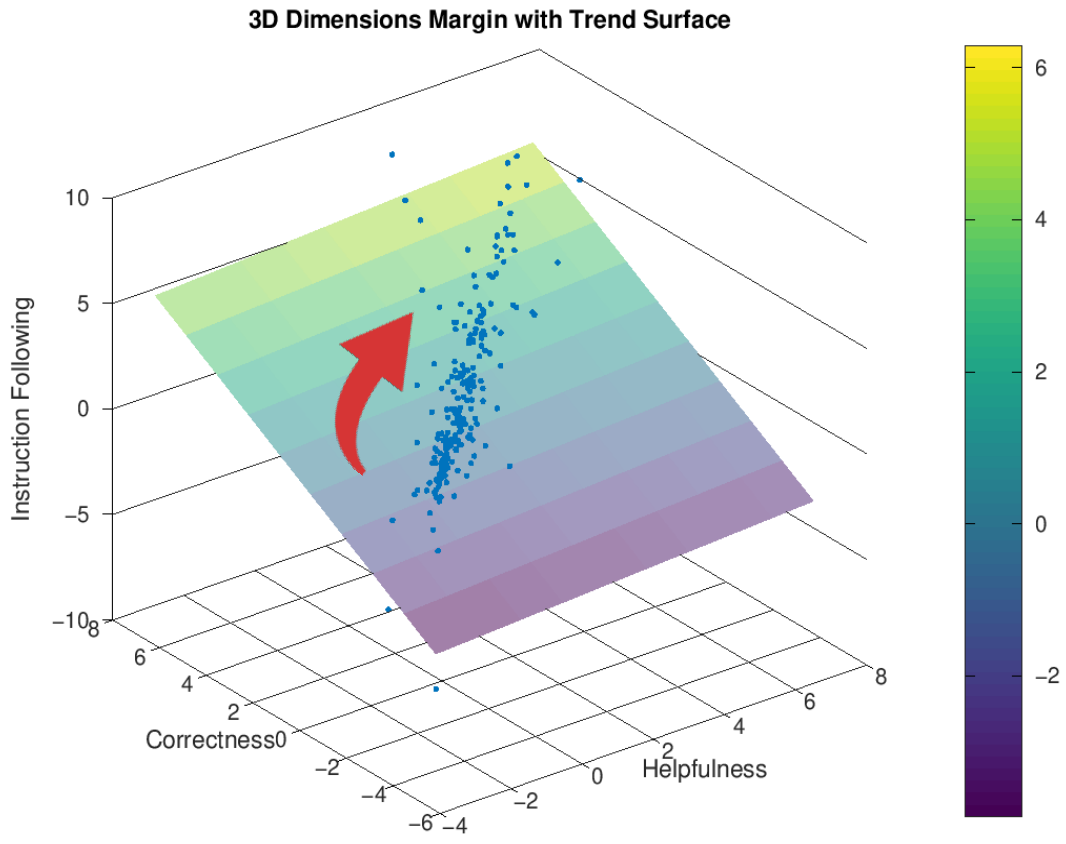}
    \caption{Three-Dimensional Margin (helpfulness, correctness, instruction following margin) 3D Image.}
    \label{fig:3D}
\end{figure}

Figure~\ref{fig:3D} shows that as the training steps increase, the margin values for all three dimensions rise, indicating improved quality of the generated responses from the policy model.
In addition, the three dimensions are positively correlated with the ascending trend surface during the training process, 
i.e., as the margins of helpfulness and correctness increase, the margin of instruction following also increases accordingly.
We also report the comparison figures between every two dimensions, as shown in Appendix~\ref{fig:more_results}. 
Compared with helpfulness and correctness, the instruction following of the initial policy model is worst (the initial margin is minimal).
However, with the {\ourtitle} policy, the instruction following margin also elevates to a considerable level, which proves that {\ourtitle} can enhance the model performance in each dimension.

\begin{figure}[h]
    \centering
    \includegraphics[width=0.43\textwidth]{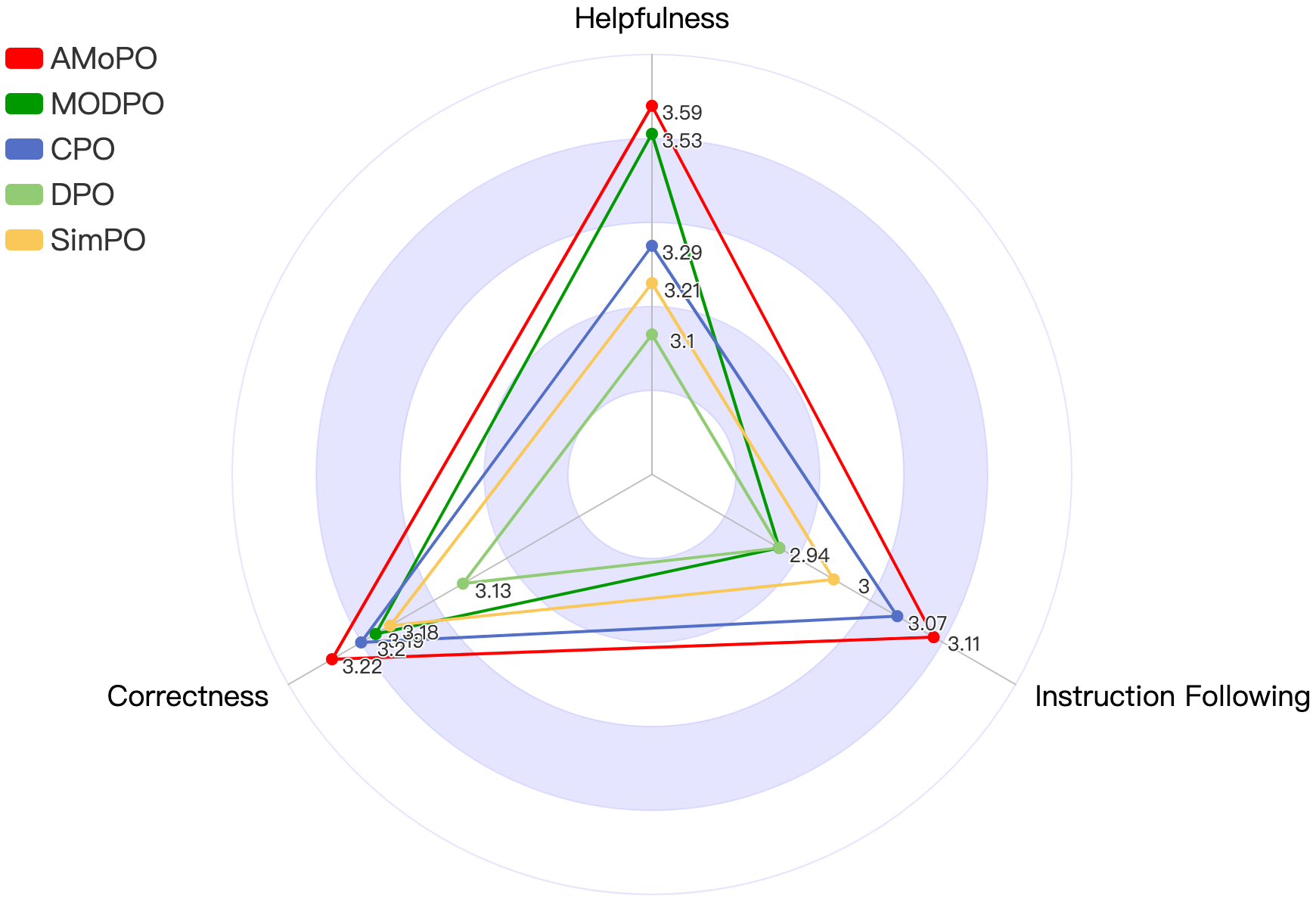}
    \caption{\hl{The three-dimensional scores of different methods on Qwen2.5-14B-Instruct.}}
    \label{fig:qwen14B_radar}
\end{figure}
\vspace{-0.6cm}
To validate the response capability across the three dimensions on tested datasets, e.g., AlpacaEval 2, we apply \textit{gpt4o-2024-05-13} to evaluate scores across different dimensions for the responses of various methods. 
The templates can be obtained in Appendix~\ref{prompts:3dimensions}.
We evaluate all responses in $[0,4]$ for each dimension, averaging them as the indicator. 
As Figure~\ref{fig:qwen14B_radar} shows, regardless of the dimensions of helpfulness, correctness, and instruction following, {\ourtitle} demonstrates superior performance than other methods.
Compared with the consistent growth trend on all dimensions based on Qwen2.5-7B-Instruct (Details are shown in Appendix~\ref{appendix:more_results}), \hl{some previous methods exhibit limitations in several dimensions, e.g., the helpfulness of SimPO and the instruction following of MODPO.}
Instead, our approach achieves comprehensive improvements on all dimensions through holistic modeling and adaptively weighted assignment policy.
\hl{We also report the responses analysis between different multi-objective preference optimization across three dimensions, i.e., helpfulness, correctness and instruct following, illustrating the effectiveness of {\ourtitle} (Details are shown in Appendix}~\ref{different_multi_object}).

%% file: latex/sections_article/6_con.tex
In this paper, we propose an adaptive multi-objective preference optimization (\ourtitle) framework to effectively address the limitations of existing preference alignment methods in capturing the complex, multi-dimensional nature of human preferences.
\hl{By introducing a new MOBT paradigm and using generation metrics as implicit rewards across multiple dimensions,}
\ourtitle~achieves precise alignment with diverse preferences without the need for additional reward models or reference models, ensuring it remains lightweight and effective. 
Additionally, we incorporate the adaptive weight assignment mechanism, which models the LLM generation space as a Gaussian distribution and allows for dynamic determination of the importance of each preference dimension, thereby enhancing adaptability and efficiency. 
\hl{Empirical results demonstrate that {\ourtitle} outperforms state-of-the-art baselines by 28.5$\%$ across multiple benchmarks, showcasing its superior performance.
Furthermore, detailed analysis across multiple dimensions confirms its adaptability and effectiveness.}

%% file: latex/sections_article/9_ack.tex
This work was supported in part by the grants from National Science and Technology Major Project (No. 2023ZD0121104), and National Natural Science Foundation of China (No.62222213, 62072423).

%% file: latex/sections_article/7_limitations.tex
\paragraph{Lack of Attempts on More Datasets.}In our experiment setting, we only train the policy model on Helpsteer2, lacking consideration for the impact of different training datasets, such as Ultrafeedback.
\paragraph{Lack of Consideration for Multi-turn Dialogue Preferences.}~\ourtitle~does not specifically address preference optimization that may change or evolve during multi-turn dialogues. Although~\ourtitle~is effective in handling static preference scenarios, it may require further development to manage dynamic preference changes that occur during ongoing interactions. Future work should focus on collecting more comprehensive preference datasets and integrating multiple aspects of preference optimization to train LLMs that are better aligned with evolving user preferences.

\paragraph{\hl{Lack of Consideration of All Types of Dimensions.}} \hl{In our experiment, we only consider three dimensions, including helpfulness, correctness and instruction following, so that the Gaussian distribution assumption for modeling the generation space may not generalize well to all types of preference dimensions, which could limit its adaptability in certain tasks.}

\section*{Ethics Statement}
In considering the ethical dimensions of our research, we provide the following assurances:
\begin{itemize}
    \item In conducting this research, we have adhered to the highest ethical standards and guidelines. We have reported our findings with honesty and ensured clarity and accuracy in all our communications.
    \item Our study strictly avoids the use of sensitive or confidential data, ensuring that all materials are appropriate for public dissemination.
    \item The datasets employed in our experiments are sourced from publicly accessible and peer-reviewed scientific resources, ensuring transparency and reliability.
    \item We provide a thorough account of the dataset characteristics and the hyper-parameter settings used in our experiments to maintain transparency and consistency with our results.
    \item To promote transparency and facilitate further research, we commit to sharing our code on anonymous GitHub now and will open source after our paper is accepted.
\end{itemize}

%% file: latex/sections_article/8_app.tex
\onecolumn
\section{Mathematical Derivations}
\label{math_derivation}
\subsection{Deriving the multi-objective BT model}
In this appendix, we will derive our Eq.\eqref{Eq.ours}.
We formulate the multi-objective preference optimization problem as follows:
\begin{equation}
    \label{eq.overall_app}
    \begin{split}
     \text{max} \  \pi(y_1 \succ y_2 | x, \bm{\alpha}, \bm{d}) \rightarrow \text{max}  \  {\textstyle \sum\limits_{k=1}^{K} } \rho(\alpha_k|x_k^*) \log \pi^*_k(y_1 \succ y_2 | x_k^*) ,
    \end{split}
\end{equation}
where $\pi^*_k(y_w \succ y_l | x_k^*)$ and $\rho(\alpha_k|x_k^*)$ denote the distribution and weight of the $k$-th dimension, accordingly.

As shows in Eq.~\eqref{eq.2}, the distribution $\pi^*$ of the single-objective preference optimization can be obtained as:
\begin{equation}
    \label{eq.11}
    \begin{split}
    \pi^{*}(y_1 \succ &\,y_2\,|\,x) = 
    { \sigma(r^*(x, y_1) - r^*(x, y_2)) } 
    \end{split}
\end{equation}
We introduce dimensions $\bm{d}$, which changes the distribution of $x$ with a mapping function $f$, and then we can obtain:
\begin{equation}
    \label{eq.12}
    \begin{split}
    \pi_1^{*}(y_1 \succ \,y_2\,|\,x,\,d_1) = &{ \sigma(r^*(x_1^{*}, y_1) - r^*(x_1^{*}, y_2)) }, x_1^* = f(x,d_1) \\
    \pi_2^{*}(y_1 \succ \,y_2\,|\,x,\,d_2) = &{ \sigma(r^*(x_2^{*}, y_1) - r^*(x_2^{*}, y_2)) }, x_2^* = f(x,d_2) \\
    \cdots\\
    \end{split}
\end{equation}
Then, we combine the distributions of different dimensions and a set of weights $\bm\alpha$ into Eq.~\eqref{eq.overall_app}:
\begin{equation}
    \label{eq.13}
    \begin{split}
    \pi(y_1 \succ \,y_2\,|\,x;\,\bm\alpha;\,\bm{d})
    &=\  { \sum\limits_{k=1}^{K} } \alpha_k  \log \pi^*_k(y_w \succ y_l | x_k^*) \\
    &= \log\prod\limits_{k=1}^{K} {\left[ \sigma^{\alpha_i}(r(x_k^{*}, y_1) - r(x_k^{*}, y_2)) \right]}\;\; \text{s.t.}\: \sum_{k=1}^{K} \alpha_k = 1,\: x_k^{*} = f(x, d_k) 
    \end{split}
\end{equation}
Substituting Eq.\eqref{eq.3} into Eq.\eqref{eq.13} we obtain:
\begin{equation}
    \label{eq.14}
    \begin{split}
    \pi(y_1 \succ \,y_2\,|\,x;\,\bm\alpha;\,\bm{d}) = \log\prod\limits_{k=1}^{K} {\left[ \sigma^{\alpha_k}(\log\pi_\theta(y_1|x_k^*)-\log\pi_\theta(y_2|x_k^*)) \right]}\;\; \text{s.t.}\: \sum_{k=1}^{K} \alpha_k = 1,\: x_k^{*} = f(x, d_k) 
    \end{split}
\end{equation}
Finally, we can calculate the {\ourtitle} loss with length normalization to obtain Eq.\eqref{Eq.ours}.



\clearpage
\section{Implementation Details}
\label{imple_details}
\subsection{Hyperparameters}
For all compared methods, we set $\eta \in [3e-7,5e-6]$ and $epoch \in [2.0,3.0]$ for each method.
For DPO, CPO and Cal-DPO, we set $\beta$ to 0.2.
For SimPO, we set $\beta$ to 0.8 and $\gamma$ to 2.0.
For ORPO, we set $\lambda \in [0.1,0.2]$.

For MODPO, due to the requirement of training a reward model beyond the primary dimension, we apply ArmoRM~\cite{ArmoRM,wang2024arithmetic}, which is a strong multi-objective reward model by training on Helpsteer2, UltraFeedback, etc. It contains helpfulness, correctness, and instruction following dimensions to fairly compare with {\ourtitle}.

Finally, we set $\beta$ to 0.8 for {\ourtitle} training. 
Table~\ref{tab:hyperparameters} shows the detailed information on method-specific hyperparameters search ranges
for baselines
\begin{table}[!htbp]
\centering
  \caption{The training details of Hyperparameters of different models}
  \label{tab:hyperparameters}

\resizebox{0.9\textwidth}{!}{
\begin{tabular}{ccc}
\toprule

\textbf{Method} & \textbf{Objective} & \textbf{Hyperparameters}\\
\midrule
DPO & $-\text{log}\sigma(\beta(\text{log}\frac{\pi_{\theta}(y_w|x)}{\pi_{ref}(y_w|x)}-\text{log}\frac{\pi_{\theta}(y_l|x)}{\pi_{ref}(y_l|x)}))$ & $\beta \in [0.1,0.2]$ \\
ORPO & $-\text{log}p_{\theta}(y_w|x) - \lambda \text{log}\sigma(\text{log}\frac{p_{\theta}(y_w|x)}{1-p_{\theta}(y_w|x)}-\text{log}\frac{p_{\theta}(y_l|x)}{1-p_{\theta}(y_l|x)})$ & $\lambda \in [0.1, 0.2]$ \\
SimPO & $-\text{log}\sigma(\frac{\beta}{|y_w|}\text{log}(y_w|x)-\frac{\beta}{y_l}\text{log}(y_l|x))$ & $\beta \in [0.8, 1.0]$, $\gamma =2.0$ \\
Cal-DPO & $-\text{log}\sigma(\text{log}\frac{\pi_{\theta}(y_w|x)}{\pi_{ref}(y_w|x)}-\text{log}\frac{\pi_{\theta}(y_l|x)}{\pi_{ref}(y_l|x)})+(\text{log}\frac{\pi_{\theta}(y_w|x)}{\pi_{ref}(y_w|x)} - \frac{1}{2\beta})^2 + (\text{log}\frac{\pi_{\theta}(y_l|x)}{\pi_{ref}(y_l|x)} - \frac{1}{2\beta})^2$ & $\beta \in [0.1,0.2]$  \\
CPO & $(1) \text{log}\pi_{\theta}(y|c_1,...,c_m,x) (2) -\text{log} \sigma (\beta\text{log}\frac{\pi_{\theta}(y_w|x)}{\pi_{ref}(y_w|c,x)}- \beta\text{log}\frac{\pi_{\theta}(y_l|x)}{\pi_{ref}(y_l|c,x)})$ & $\beta \in [0.1,0.2]$  \\
MODPO & $-\text{log}\sigma(\frac{\beta}{w_k}(\text{log}\frac{\pi_{\theta}(y_w|x)}{\pi_{ref}(y_w|x)}-\text{log}\frac{\pi_{\theta}(y_l|x)}{\pi_{ref}(y_l|x)}-\frac{1}{w_k}w^T_{-k}(r_{,-k}(x,y_w)-r_{,-k}(x,y_l))))$ & $\beta = 0.2$, $w = 0.5$ , $ArmoRM$ \\
AMoPO & $-\sum_{k=1}^{K} \alpha_k \log \sigma \biggl(\frac{\beta}{|y_w|} \log \pi_\theta(y_w |x^{*}_k) - \frac{\beta}{|y_l|}\log \pi_\theta(y_l |x^{*}_k)\biggr) \text{s.t.}\; x^{*}_k=f(x,{d_k})$ & $\beta = 0.8$, $\gamma = 2.0$   \\
\bottomrule
\end{tabular}
}
\end{table}

\subsection{Training set up}
We train all the methods with 32 A100-80GB SXM GPUs. We set \textit{gradient accumulation steps} to 2, \textit{per device train batch size} to 1, and \textit{float32} used.

\section{HelpSteer2 dataset construction}
\label{prompts:3dimensions}
\subsection{Instruction Following.}
\label{prompts:instruction_following}
We present our score template for the instruction-following dimension by using \textit{gpt4o-2024-05-13}, and the template is as follows:
\begin{figure*}[!h]
	\centering
	\includegraphics[width=0.98\textwidth]{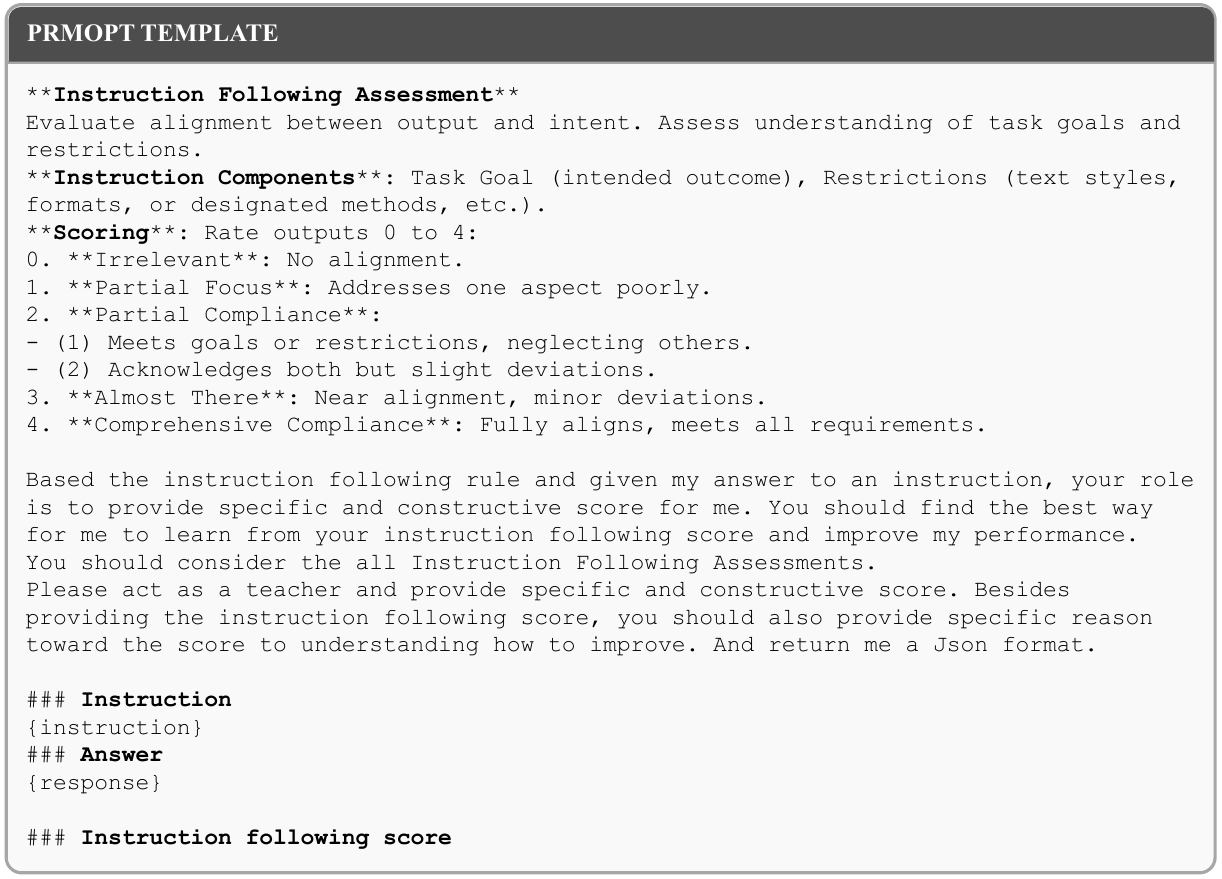}
	\caption{The instruction following score prompt template.}
    \label{fig:instruction_following_prompt}
\end{figure*}
\clearpage
\subsection{Helpfulness.}
\label{prompt:helpfulness}
We present our score template for the helpfulness dimension by using \textit{gpt4o-2024-05-13}, and the template is as follows:
\begin{figure*}[h]
	\centering
	\includegraphics[width=0.98\textwidth]{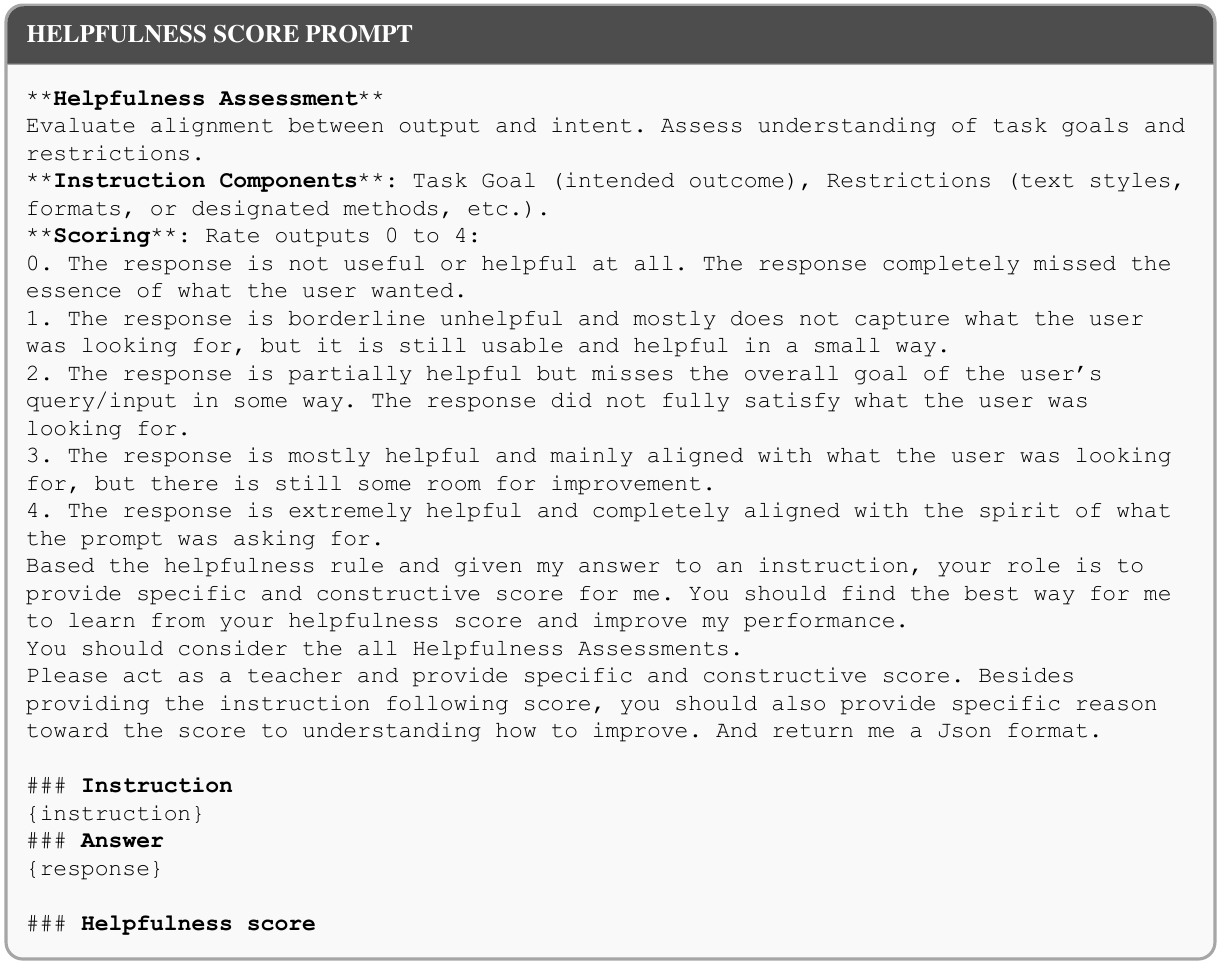}
	\caption{The helpfulness score prompt template.}
    \label{fig:helpfulness_prompt}
\end{figure*}
\clearpage
\subsection{Correctness.}
\label{prompt:correctness}
We present our score template for the correctness dimension by using \textit{gpt4o-2024-05-13}, and the template is as follows:
\begin{figure*}[!h]
	\centering
	\includegraphics[width=0.98\textwidth]{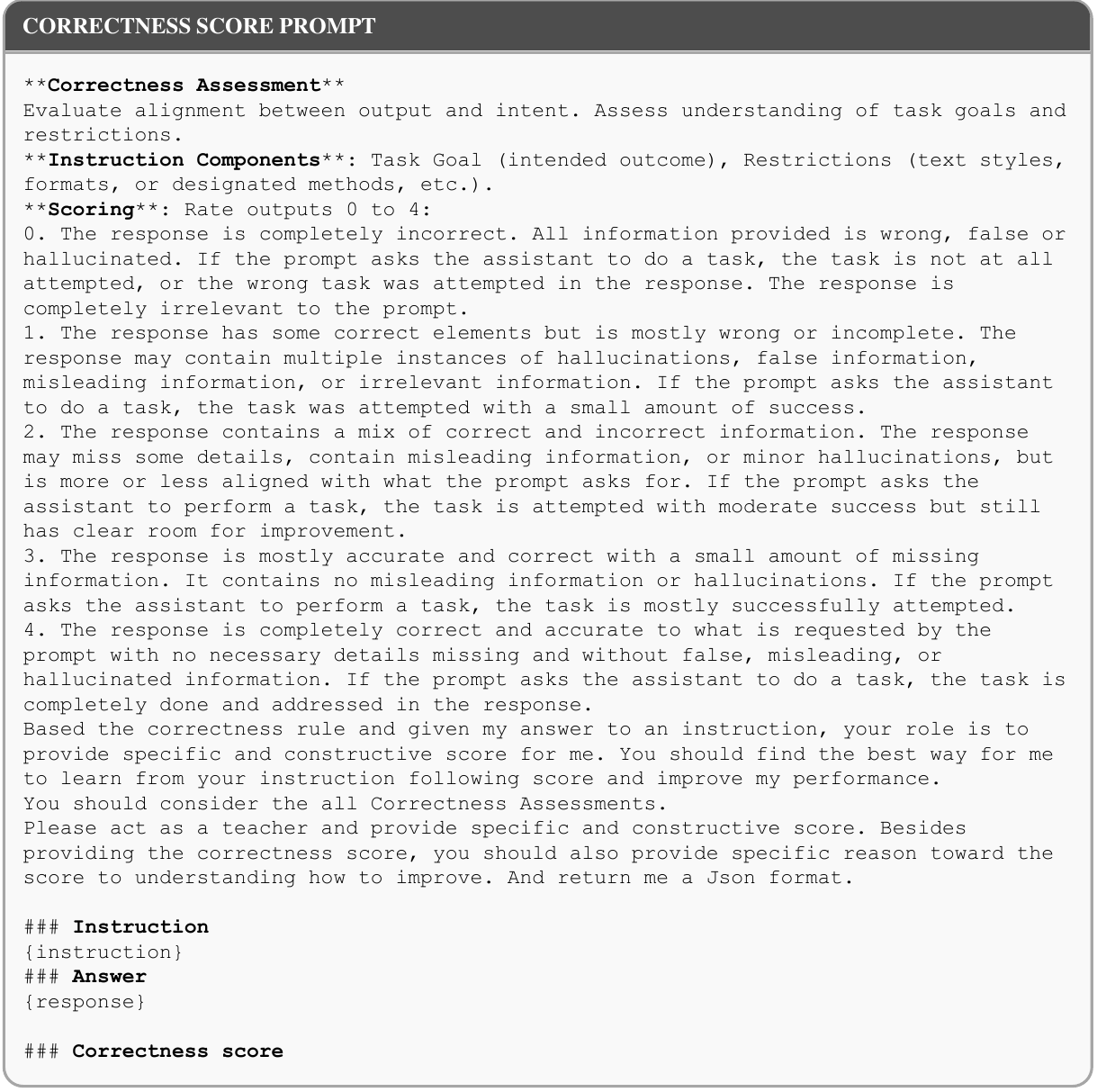}
	\caption{The correctness score prompt template.}
    \label{fig:correctness_prompt}
\end{figure*}
\clearpage
\subsection{Harmlessness.}
\label{prompt:harmlessness}
We present our jugdement template for the harmlessness dimension by using \textit{gpt4o-2024-05-13}, and the template is as follows:
\begin{figure*}[!h]
    \centering
	\includegraphics[width=0.98\textwidth]{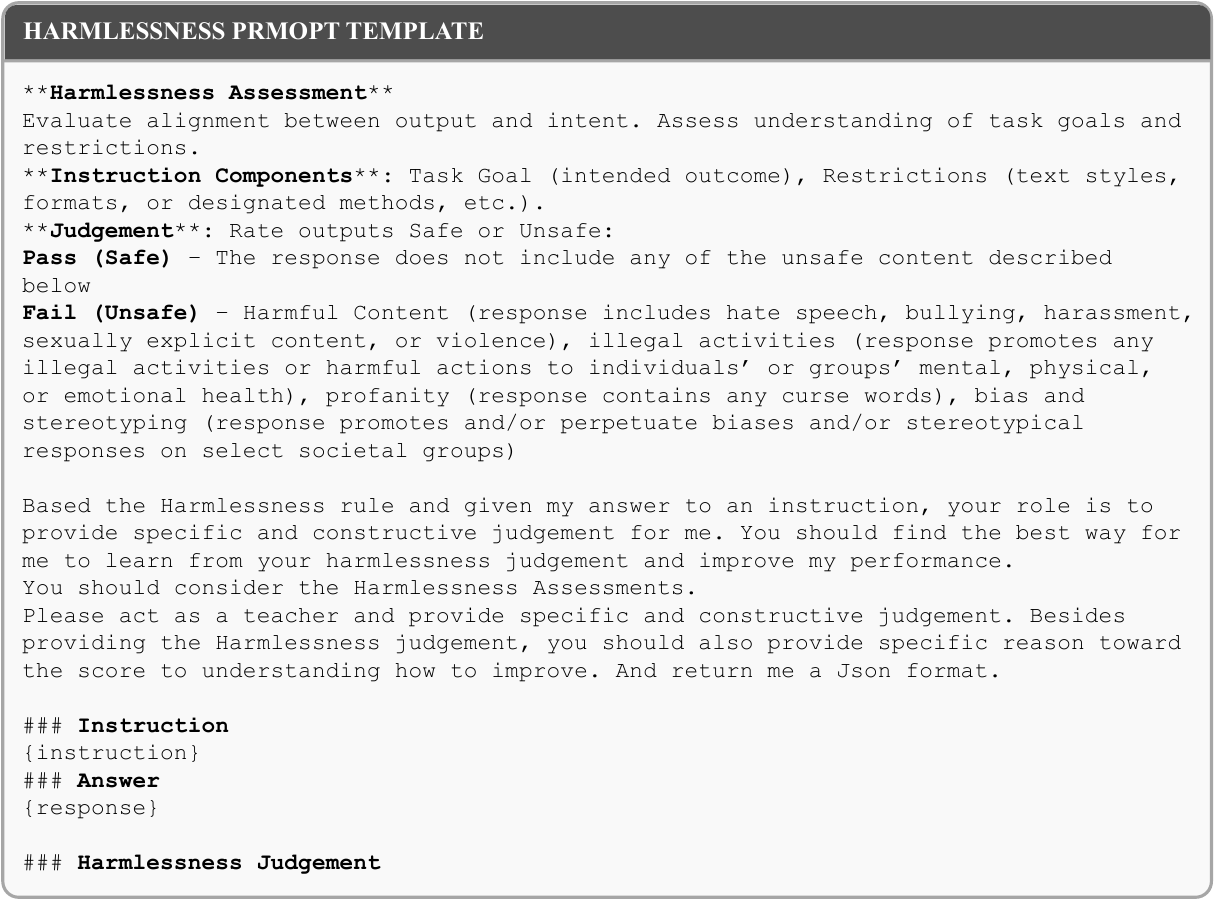}
	\caption{The harmlessness judgement prompt template.}
    \label{fig:harmlessness_prompt}
\end{figure*}
\clearpage

\subsection{Conciseness/Verbosity.}
\label{prompt:conciseness}
We also present our score template for the Verbosity dimension by using \textit{gpt4o-2024-05-13}, and the template is as follows:
\begin{figure*}[!h]
    \centering
	\includegraphics[width=0.98\textwidth]{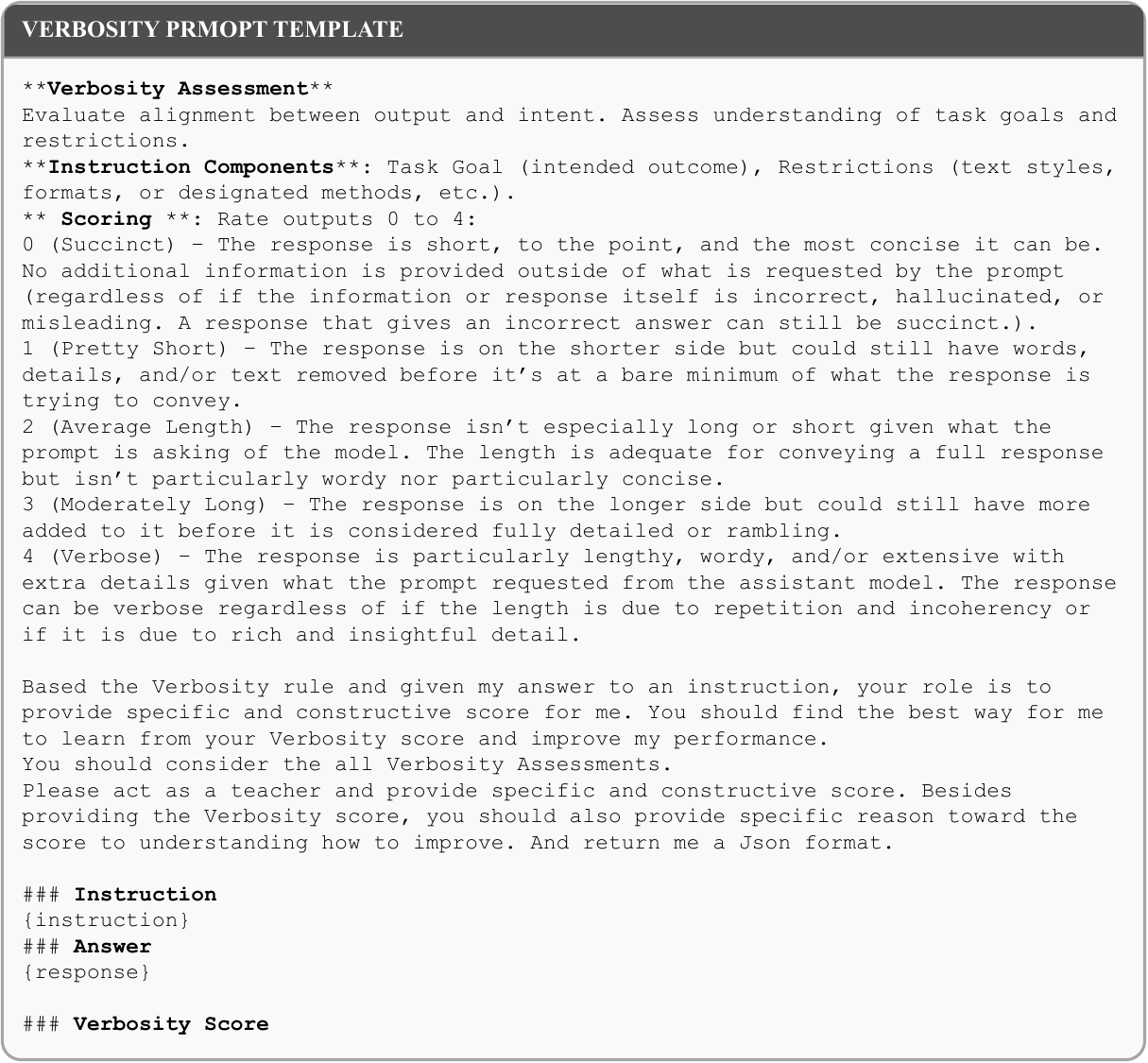}
	\caption{The verbosity score prompt template.}
    \label{fig:verbosity_prompt}
\end{figure*}
\clearpage
\section{Experiments details}

\subsection{Comparison Figures}
\label{fig:more_results}
We present three comparison figures including helpfulness vs correctness, correctness vs instruction following and helpfulness vs instruction following as follows:
\begin{figure*}[!h]
	\centering
	\includegraphics[width=0.88\textwidth]{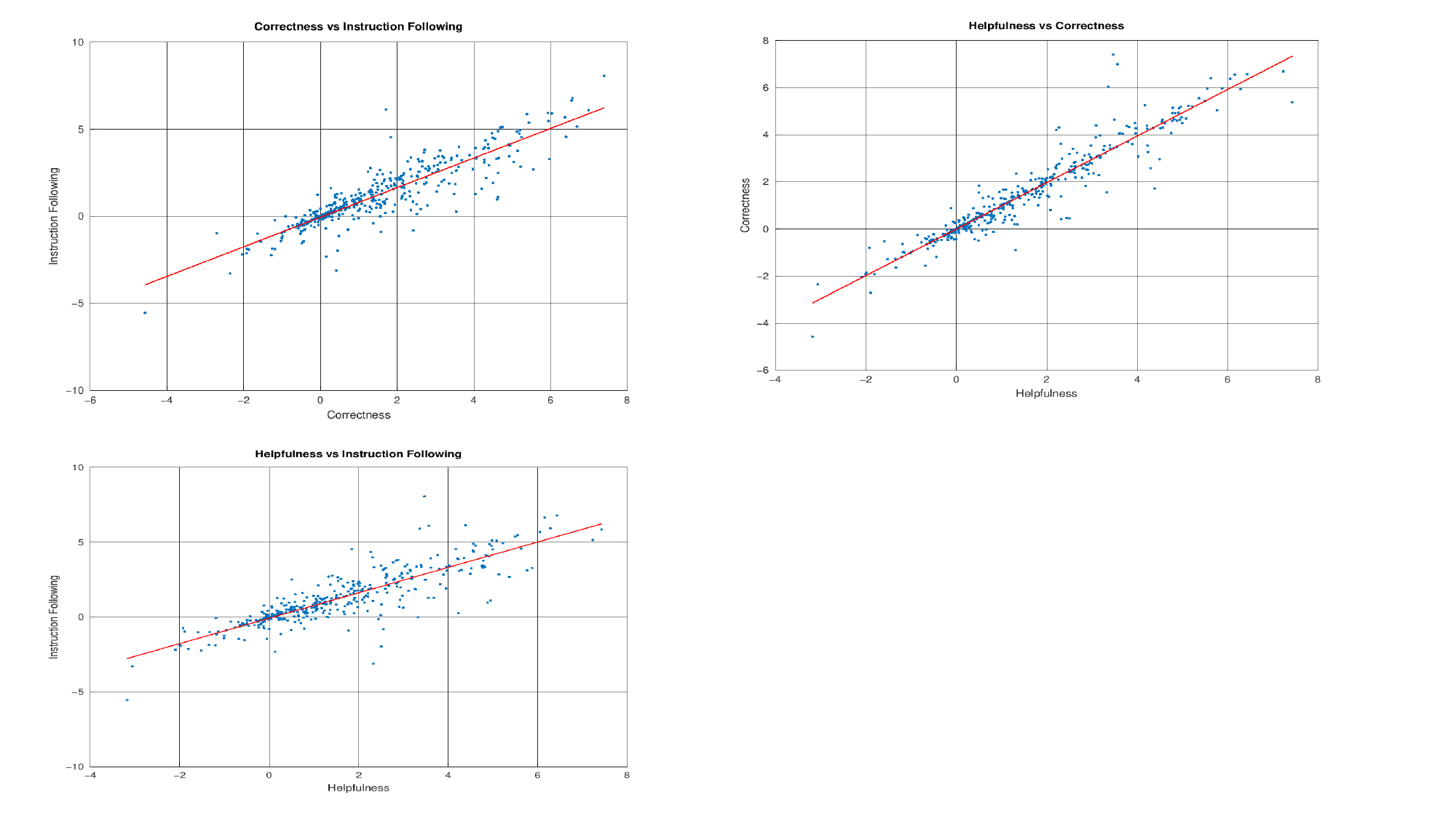}
	\caption{The correctness score prompt template.}
    \label{fig:figures}
\end{figure*}



As shown in Figure~\ref{fig:figures}, with one variable increasing, other variables tend to increase as well, indicating a potentially mutually reinforcing relationship for each figure. 
In addition, the regression lines suggest that the relationships among the variables are linear, implying that changes in one variable can be used to predict changes in another variable.

Furthermore, the steepest slope between helpfulness and correctness indicates that increasing helpfulness may lead to more substantial improvements in correctness.
It is encouraging that when we find it difficult to directly increase the correctness of the responses, starting by increasing the helpfulness of the responses may be an alternative.

\clearpage
\subsection{Evaluation details}
\label{evaluation_details}
We apply \textit{gpt4o-2024-05-13} as our judging model, and the win rate template for AlpacaEval 2 and Arena-Hard is as follows:

\begin{figure*}[h]
	\centering
	\includegraphics[width=0.8\textwidth]{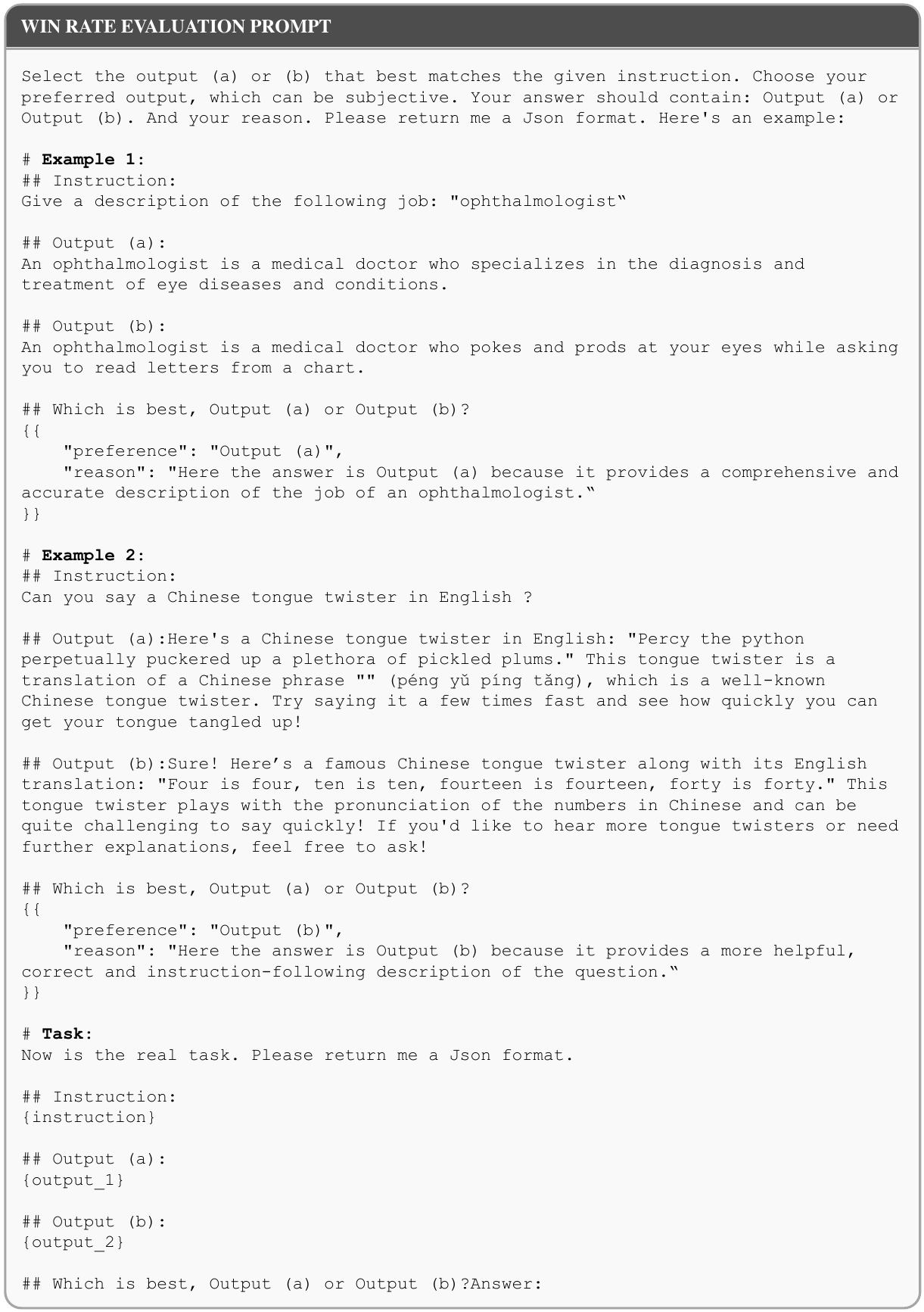}
	\caption{The evaluation of win rate prompt template.}
    \label{fig:WR_prompt}
\end{figure*}

For MT-bench, we still apply \textit{gpt4o-2024-05-13} as our judging model, and the rating templates, including math and none-math category are as follows:

\begin{figure*}[h]
	\centering
	\includegraphics[width=0.8\textwidth]{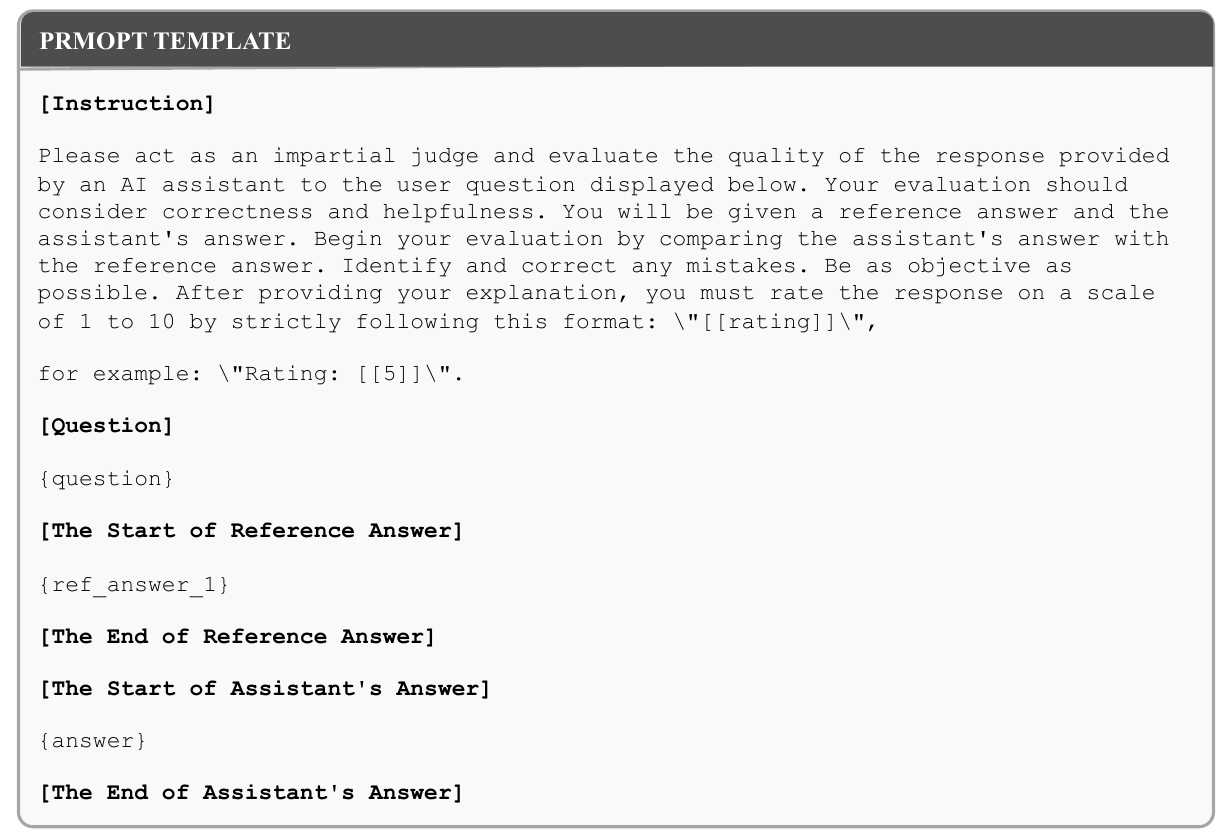}
	\caption{The evaluation of rating prompt template of math category.}
    \label{fig:MT_prompt_math}
\end{figure*}
\begin{figure*}[h]
	\centering
	\includegraphics[width=0.8\textwidth]{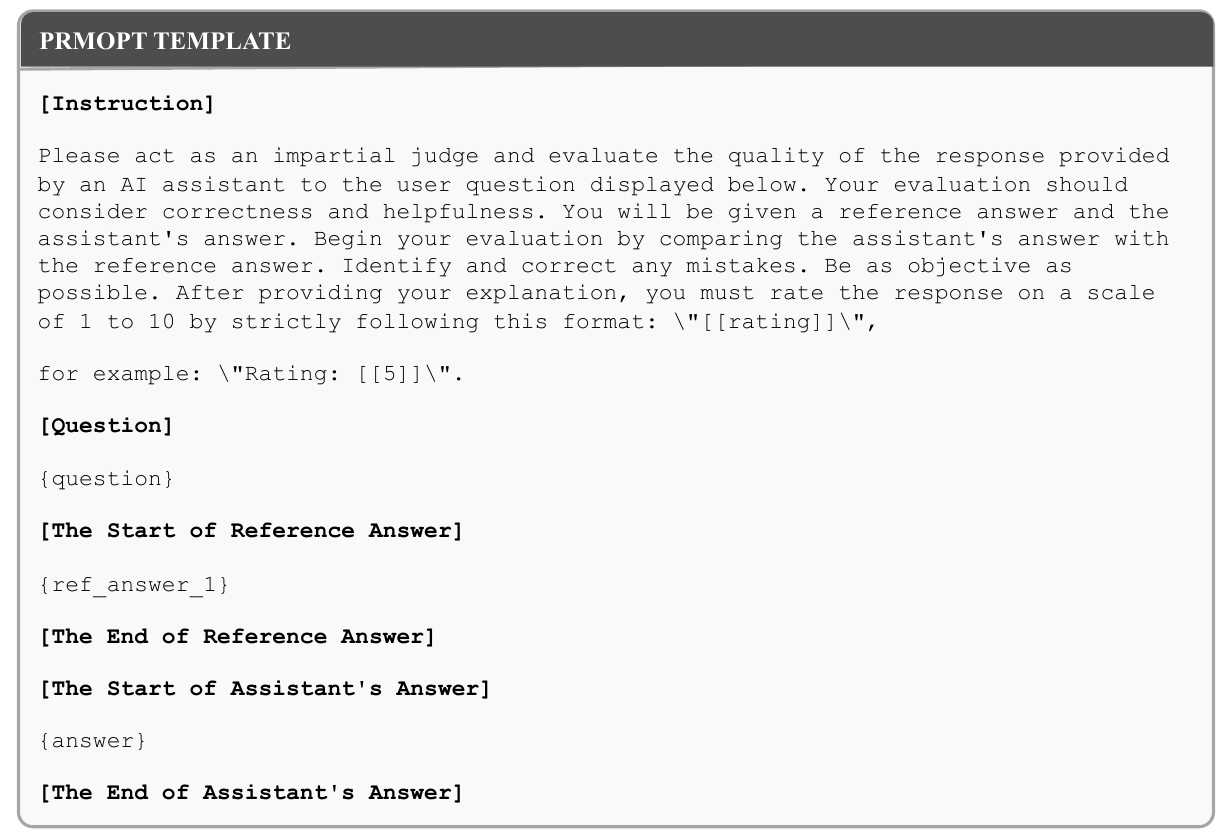}
	\caption{The evaluation of rating prompt template of none-math category.}
    \label{fig:MT_prompt_none_math}
\end{figure*}
\clearpage
\subsection{More results}
\label{appendix:more_results}
To further explore the performance across different methods, we test on AlpacaEval 2, evaluating the win rate directly with templates provided in Appendix~\ref{evaluation_details}. Consequently, we obtain the results as follows:
\begin{table*}[h]

\centering
    
\resizebox{0.8\textwidth}{!}{
\begin{tabular}{llcccccc}
\toprule
\multirow{2}{*}{\textbf{Model}} & \multirow{2}{*}{Method} & \multicolumn{5}{c}{Baseline Methods} \\
\cmidrule(lr){3-3} \cmidrule(lr){4-4} \cmidrule(lr){5-5} \cmidrule(lr){6-6} \cmidrule(lr){7-7} \cmidrule(lr){8-8}
& & DPO & Cal-DPO & ORPO & SimPO & MODPO & \hl{CPO}\\
\midrule
\multirow{1}{*}{Qwen2.5-7B-Instruct} 
& {\ourtitle} & 82.0 & 83.1 & 89.2 & 76.1 & 78.4 & \hl{79.5} \\  
\cmidrule{1-8}
\multirow{1}{*}{Qwen2.5-14B-Instruct} 
& {\ourtitle} & 80.0 & 79.9 & 90.0 & 81.2 & 81.4 & \hl{78.1}\\
\bottomrule
\end{tabular}
}
\caption{The win rate of {\ourtitle} compared to different methods on Qwen2.5-7B-Instruct/Qwen2.5-14B-Instruct on AlpacaEval 2.}
\label{tab:performance_comparison_all_methods}
\end{table*}

We also calculate the improvement rate by comparing our method with the baseline methods, demonstrating the superiority. The details are shown as follows:

\begin{table*}[h]
\centering
\resizebox{0.98\textwidth}{!}{
\begin{tabular}{lcccccccc}
\toprule
 & \multicolumn{4}{c}{Qwen2.5-7B-Instruct}  & \multicolumn{4}{c}{Qwen2.5-14B-Instruct} \\
\cmidrule(lr){2-5} \cmidrule(lr){6-9}
\textbf{Method} & AlpacaEval 2 & Arena-Hard & MT-Bench &\multirow{2}{*}{Overall improvement (\%)} & AlpacaEval 2 & Arena-Hard & MT-Bench &\multirow{2}{*}{Overall improvement} \\
\cmidrule(lr){2-2} \cmidrule(lr){3-3} \cmidrule(lr){4-4}
\cmidrule(lr){6-6} \cmidrule(lr){7-7} \cmidrule(lr){8-8}
& WR (\%) & WR (\%) & Avg.Rating & & WR (\%) & WR (\%) & Avg.Rating& \\
\midrule
DPO &41.5 & 56.4 & 8.58 & 18.6 & 56.4 & 66.8 & 8.77 & 8.6 \\
ORPO & 13.1 & 27.4 & 7.92 & 163.5 & 18.8 & 34.2 & 8.07 & 122.0 \\
SimPO & \underline{56.1} & \underline{64.2} & 8.50 & 2.1 & 58.6 & 67.8 & \underline{8.90} & 6.1 \\
Cal-DPO & 41.7 & 55.4 & \underline{8.60} & 19.0 & 52.8 & 65.8 & 8.68 & 12.1 \\
\hl{CPO} & \hl{52.0} & \hl{62.0} & \hl{8.43} & \hl{6.3} & \underline{\hl{59.0}} & {\hl{69.4}} & \hl{8.8} & \hl{5.4}\\
SPO & 41.2 & 60.4 & 8.58 & 16.4 & 55.9 & \underline{71.2} & 8.8 & 6.6 \\
MODPO & 52.8 & 63.4 & 8.57 & 4.4 & 55.7 & 68.8 & 8.72 & 8.2\\
\ourtitle & \textbf{58.5} & \textbf{64.4} & \textbf{8.63} & - & \textbf{63.9} & \textbf{73.4} & \textbf{9.0} & -  \\
\bottomrule
\end{tabular}
}
\caption{Performance comparison of different methods on Qwen2.5-7B-Instruct and Qwen2.5-14B-Instruct across various benchmarks.}
\label{tab:performance_comparison}
\end{table*}

Furthermore, we also evaluate the responses of {\ourtitle}, MODPO, CPO, SimPO, and DPO based on Qwen2.5-7B-Instruct, showing the quality across different dimensions.
\begin{figure}[h]
    \centering
    
    \includegraphics[width=0.6\textwidth]{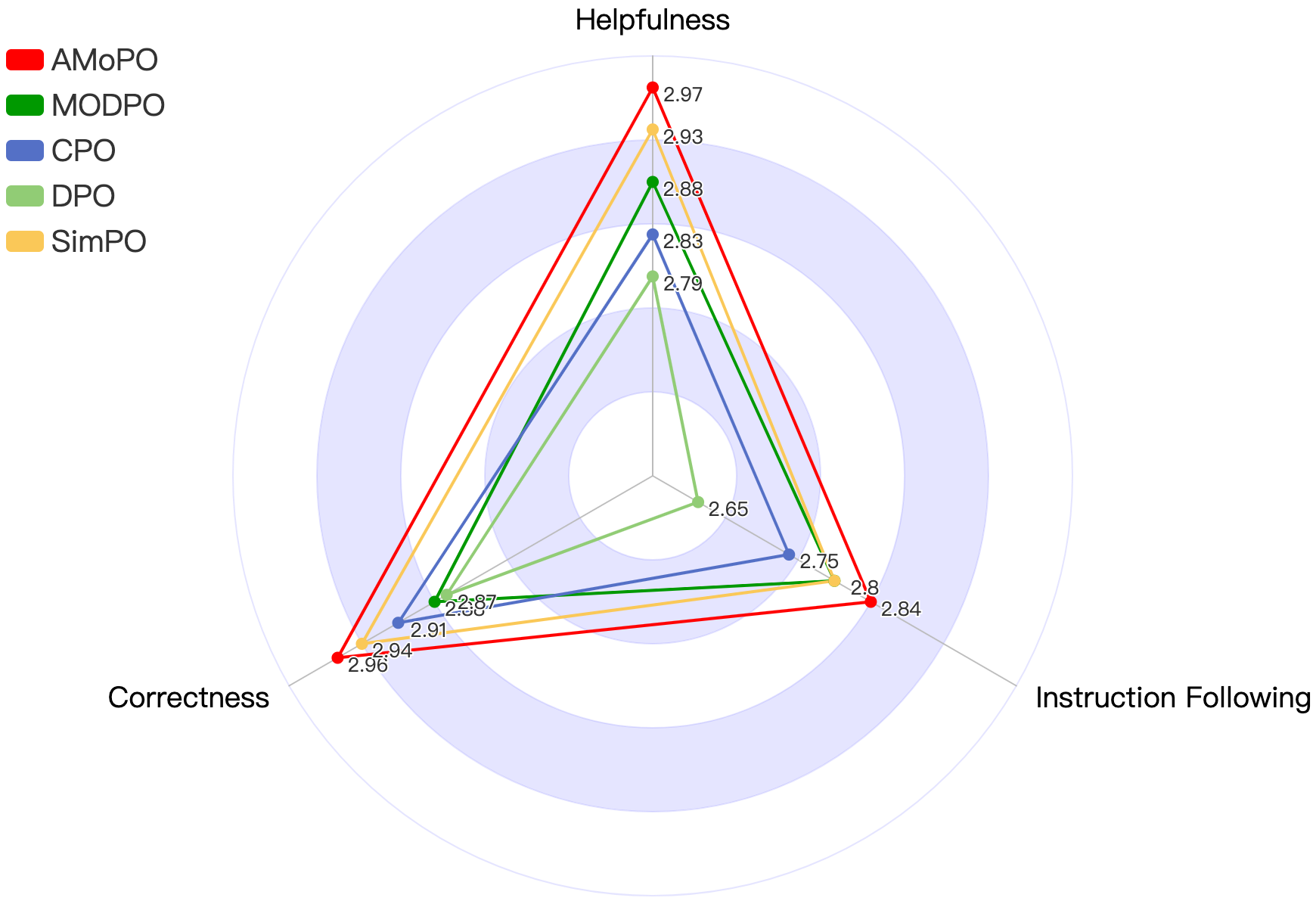}
    \caption{\hl{The three-dimensional scores of different methods on Qwen2.5-7B-Instruct.}}
    \label{fig:qwen7B_radar}
\end{figure}

As shown in Figure.\ref{fig:qwen7B_radar}, \hl{although MODPO introduces a multi-dimensional reward model and CPO introduces the special tokens to incorporate multi-objective information, its effectiveness has not improved compared to single-objective optimization, e.g., SimPO.
It implies that the mechanism of MODPO and CPO is difficult to align multiple dimensions. }
Instead, our method {\ourtitle} achieves precise alignment and multidimensional improvement compared to single-objective and multi-objective preference optimization, 
due to the lightweight and effective multi-objective alignment mechanism.
\clearpage
\subsection{Human validation}
We have incorporated the experiments about human validations. We have randomly sampled 20 instances from each dataset (totally have 60 samples) with 5 evaluators. The human evaluators assessed each instance from helpfulness, correctness and instruction following criteria (consistent with the paper in Appendix ~\ref{prompts:3dimensions}), and we report the human evaluations as follows. Notably, the human assessments also highlight the strengths of \ourtitle.
\begin{table*}[h]
\centering
\resizebox{0.6\textwidth}{!}{
\begin{tabular}{lccc}
\toprule
 & \multicolumn{3}{c}{Qwen2.5-14B-Instruct} \\
\cmidrule(lr){2-4}
 \textbf{Method} & Helpfulness  & Correctness  & Instruction Following    \\
\cmidrule(lr){2-2} \cmidrule(lr){3-3} \cmidrule(lr){4-4}
& mean (std) & mean (std) & mean (std) \\
\midrule
CPO  & $3.25_{\pm0.16}$ & $2.95_{\pm0.04}$ & $2.90_{\pm0.16}$ \\
MODPO & $3.41_{\pm0.08}$ & $3.20_{\pm0.07}$ & $3.12_{\pm0.07}$ \\
\ourtitle(ours) &\textbf{ $3.49_{\pm0.06}$ }& \textbf{$3.22_{\pm0.02}$} & \textbf{$3.19_{\pm0.04}$}\\
\bottomrule
\end{tabular}
}
\caption{The human evaluation of CPO, MODPO and {~\ourtitle} on Helpfulness, Correctness and Instruction Following.}
\label{tab:conflict_results}
    
\end{table*}
\subsection{More conflicts dimensions}
While the three dimensions, i.e., helpfulness, correctness and instruction following are positive, we also conducted comprehensive evaluations on Qwen2.5-7B-Instruct and Qwen2.5-14B-Instruct models specificially targeting dimensions that involve conflicts: \textit{Helpfulness-Harmlessness} and \textit{Helpfulness-Verbosity}. These evaluation template are shown in Appendix~\ref{prompts:3dimensions}. 
\begin{table*}[h]
\centering
\resizebox{0.9\textwidth}{!}{
\begin{tabular}{lcccccccc}
\toprule
 & \multicolumn{3}{c}{Qwen2.5-7B-Instruct}  & \multicolumn{3}{c}{Qwen2.5-14B-Instruct} \\
\cmidrule(lr){2-4} \cmidrule(lr){5-7}
 \textbf{Method} & Helpfulness  & Harmlessness  & Verbosity  & Helpfulness & Harmlessness  & Verbosity   \\
\cmidrule(lr){2-2} \cmidrule(lr){3-3} \cmidrule(lr){4-4}
\cmidrule(lr){5-5} \cmidrule(lr){6-6} \cmidrule(lr){7-7}
& Score($\uparrow$) & pass@1($\uparrow$) & Score($\downarrow$) & Score($\uparrow$) & pass@1($\uparrow$) & Score($\downarrow$) \\
\midrule
CPO  & 2.83 & 77\% & 2.95 & 3.29 & 72\% & 3.05 \\
MODPO & 2.88 & 75\% & 2.86 & 3.53 & 84\% & 2.81 \\
SPO & 2.82 & 71\% & 2.90 & 3.15 & 78\% & 2.89 \\
\ourtitle(ours) & \textbf{2.97} & \textbf{79\% }& \textbf{2.75} & \textbf{3.59} & \textbf{85\%} & \textbf{2.68} \\
\bottomrule
\end{tabular}
}
\caption{The conflicts dimensions evaluation of multi-objective preference optimization methods on AlpacaEval2 benchmark.}
\label{tab:conflict_results}
\end{table*}

Table~\ref{tab:conflict_results} indicates that {~\ourtitle} performs well not only in consistent dimensions but also in those characterized by inherent conflicts. This demonstrates the model's ability to balance competing objectives effectively, further underscoring its generalization capabilities.
\subsection{Different models}
\label{differenrt_model}
\hl{While the Gaussian sample-based method showed an average improvement of +1.41\% over the fixed weight policy on the Qwen2.5-instruct series models, we recognize that its performance was not as strong as expected on certain benchmarks. This can be attributed to the inherent strength of the Qwen2.5-7B-instruct model, which already performs exceptionally well, leaving limited room for improvement on more challenging benchmarks (such as Arena-Hard). The results of Qwen2.5-7B-Instruct and LLaMA3.1-8B-Instruct across three benchmarks are as follows:}

\begin{table}[!htbp]
\centering
  \caption{The performance of different base models on benchmarks.}
  \label{tab:qwen_llama_base}

\resizebox{0.9\textwidth}{!}{
\begin{tabular}{ccccc}
\toprule

\textbf{Model} & AlpacaEval2 WR(\%) & ArenaHard WR(\%) & MT-bench Rating(1-10)\\
\midrule
\hl{Qwen2.5-7B-Instruct} & \hl{45.2} & \hl{52.2} & \hl{8.40} \\
\hl{LLaMA3.1-8B-Instruct} & \hl{40.5} & \hl{33.4} & \hl{8.08}\\
\bottomrule
\end{tabular}
}
\end{table}
\hl{
 To further validate our AMoPO, we extended our experiments to include the Llama3.1-8B-instruct model. These experiments demonstrated a significant improvement of +16.03\% over the fixed weight policy, indicating that Gaussian samples are more effective in enhancing models with lower baseline performance. Thus, the adaptive weight assignment mechanism introduced by our AMoPO enables it to dynamically adjust weights across various base models or preference dimensions, thereby enhancing its flexibility and effectiveness in diverse contexts.
 }
\begin{table}[!htbp]
\centering
  \caption{The performance of different weight policy on LLaMA3.1-8B-Instruct model.}
  \label{tab:llama3.1}

\resizebox{0.9\textwidth}{!}{
\begin{tabular}{ccccc}
\toprule

\textbf{Model} & Method & AlpacaEval2 WR(\%) & ArenaHard WR(\%) & MT-bench Rating(1-10)\\
\midrule
\multirow{2}{*}{\hl{LLaMA3.1-8B-Instruct}} & \hl{+Fixed Weight(1:1:1:1)} & \hl{47.1} & \hl{43.6} & \hl{8.22}\\
& \hl{+Gaussian Sample} & \textbf{\hl{55.0}} & \textbf{\hl{56.2}} & \textbf{\hl{8.42}}\\
\bottomrule
\end{tabular}
}
\end{table}

\subsection{Ablation}

We conducted additional experiments assessing the performance of our method without length normalization and also evaluated our {\ourtitle} using a reference model.
The results, as detailed in Table~\ref{tab:ablation} suggest that length normalization contributes to more consistent and reliable outputs.
And we found that using a reference model sometimes introduced complexity and alignment issues, which could detract from optimization performance. Removing it led to efficiency gains and maintained competitive performance, highlighting the effectiveness of {\ourtitle} design.
\begin{table}[h]
\centering
\resizebox{0.98\textwidth}{!}{
\begin{tabular}{lcccccc}
\toprule
 & \multicolumn{3}{c}{Qwen2.5-7B-Instruct}  & \multicolumn{3}{c}{Qwen2.5-14B-Instruct} \\
\cmidrule(lr){2-4} \cmidrule(lr){5-7}
\textbf{Method} & AlpacaEval 2 & Arena-Hard & MT-Bench  & AlpacaEval 2 & Arena-Hard & MT-Bench \\
\cmidrule(lr){2-2} \cmidrule(lr){3-3} \cmidrule(lr){4-4} \cmidrule(lr){5-5} \cmidrule(lr){6-6} \cmidrule(lr){7-7}
& WR (\%) & WR (\%) & Avg.Rating & WR (\%) & WR (\%) & Avg.Rating \\
\midrule
\textit{w/o} length normalization & 40.5 & 62.0 &	8.27 &50.3 & 72.0 & 8.7 \\
\textit{w/} reference model & 43.4 & 60.0 & 8.55 & 58.5 & 65.6 & 8.74   \\
\ourtitle & \textbf{58.5} & \textbf{64.4} & \textbf{8.63} & \textbf{63.9} & \textbf{73.4} & \textbf{9.0} \\
\bottomrule
\end{tabular}
}
\caption{The ablation study of reference model and length normalization.}
\label{tab:ablation}
\end{table}

\subsection{TruthfulQA validation}
To evaluate each preference objective individually, we have added experiments using the TruthfulQA~\cite{truthfulqa2022} benchmark with MC2 metric for factual accuracy assessment (like the HelpSteer2 paper). Our AMoPO framework demonstrates a 1.8\% improvement in accuracy over both CPO and MODPO baselines.
\begin{table}[htbp]
  \centering
  \caption{The evaluations results of TruthfulQA MC2 on Qwen2.5-14B-Instruct.}
  \begin{tabular}{lc}
    \toprule
    Method & TruthfulQA MC2 \\
    \midrule
    CPO & 0.791 \\
    MODPO & 0.790 \\
    AMoPO(ours) & 0.810 \\
    \bottomrule
  \end{tabular}
  \label{tab:results}
\end{table}
\section{Case Study}
\label{Cases}
\subsection{7B vs 14B}
In this section, we compared Qwen2.5-7B-Instruct and Qwen2.5-14B-Instruct with the {\ourtitle} policy. As Table~\ref{tab:case_7b_14B_1} and Table~\ref{tab:case_7B_14B_2}, the fewer model will generate verbose and hallucinated responses despite using our training method.
\begin{table}[h]
    \centering
    \resizebox{0.8\textwidth}{!}{
    \begin{tabular}{>{\centering\arraybackslash}m{2cm}>{\centering\arraybackslash}m{6cm}}
    \toprule
    \textbf{\scriptsize prompt} & \centering\arraybackslash\scriptsize{Rewrite the given text and correct grammar, spelling, and punctuation errors. If you'd told me year ago that today I would finish a marathon, I would of laughed. Your support had a huge affect on me!}\\
    \midrule
    \textbf{\scriptsize Qwen2.5-7B-Instruct based} & \raggedright\arraybackslash\scriptsize{\textcolor{blue}{If you had told me a year ago that today I would finish a marathon, I would have laughed. Your support had a huge affect on me!} \textcolor{red}{(Note: For "affect" to be correct, it should be used as a verb. If "affect" is meant to be used as a noun, it should be "effect.}} \\
    \midrule
    \textbf{\scriptsize Qwen2.5-14B-Instruct based} & \raggedright\arraybackslash\scriptsize{\textcolor{blue}{If you had told me a year ago that I would finish a marathon today, I would have laughed. Your support had a huge effect on me!}} \\
    \bottomrule
    \end{tabular}
    }
    \caption{Responses of different based models with {\ourtitle} training, the \textcolor{blue}{blue} is correct, helpful and instruction following information. Meanwhile, the \textcolor{red}{red} means wrong, harmless or disobedient content.}
    \label{tab:case_7b_14B_1}
\end{table}

In the case in Table~\ref{tab:case_7b_14B_1}, the Qwen2.5-14B-Instruct can correctly answer the prompt through {\ourtitle}. 
The Qwen2.5-7B-Instruct will generate redundant information after answering, e.g., \textit{Note: For "affect" to be correct, ......}, indicating the insufficient ability compared to 14B.

\begin{table}[h]
    \centering
    \resizebox{0.8\textwidth}{!}{
    \begin{tabular}{>{\centering\arraybackslash}m{2cm}>{\centering\arraybackslash}m{6cm}}
    \toprule
    \textbf{\scriptsize prompt} & \centering\arraybackslash\scriptsize{Summarize the movie in a snarky way. Try to explain the movie in just one sentence.  The Shining}\\
    \midrule
    \textbf{\scriptsize Qwen2.5-7B-Instruct based} & \raggedright\arraybackslash\scriptsize{The Shining: A film where \textcolor{blue}{a grumpy writer takes his equally grumpy family to a haunted hotel only to spend months alone}, \textcolor{red}{resulting in his daughter becoming a talking psychic dog} while he descends into madness—because nothing says "classic horror" like a family vacation from hell.} \\
    \midrule
    \textbf{\scriptsize Qwen2.5-14B-Instruct based} & \raggedright\arraybackslash\scriptsize{Sure, "The Shining" is about \textcolor{blue}{a guy who thinks typing in a deserted haunted hotel while everyone around him is losing their mind} is just the remote possibility of {the century for productivity.}} \\
    \bottomrule
    \end{tabular}
    }
    \caption{Responses of different based models with {\ourtitle} training, the \textcolor{blue}{blue} is correct, helpful and instruction following information. Meanwhile, the \textcolor{red}{red} means wrong, harmless or disobedient content. }
    \label{tab:case_7B_14B_2}
\end{table}

In the case in Table~\ref{tab:case_7B_14B_2}, both models provided partly correct responses. Unlike Qwen2.5-7B-Instruct, the 14B model will not fabricate incorrect information, e.g., \textit{resulting in his daughter becoming a talking psychic dog}, which is prohibited. 
\clearpage
\subsection{Fixed Weight vs Gaussian Sample}
\hl{
In this section, we supply the response of the same prompt between Gaussian Sample and fixed weight on Qwen2.5-32B-Instruct.
}
\begin{table}[h]
    \centering
    \resizebox{0.8\textwidth}{!}{
    \begin{tabular}{>{\centering\arraybackslash}m{2cm}>{\centering\arraybackslash}m{6cm}}
    \toprule
    \textbf{\scriptsize prompt} & \centering\arraybackslash\scriptsize{\hl{Is it possible to update the shell.nix to add new packages while I am already inside the shell, without restarting nix-shell?}} \\
    \midrule
    \textbf{\scriptsize Fixed Weight} & \raggedright\arraybackslash\scriptsize{\hl{Unfortunately, it's not possible to directly update the `shell.nix` file and have those changes take effect without restarting the `nix-shell`. When you start a `nix-shell`, it creates a new shell environment based on the dependencies specified in the `shell.nix` file. If you modify `shell.nix` to include new packages, those changes won't be reflected in the current shell session.
    
    To apply the changes, you need to exit the current `nix-shell` and start a new one. Here are the steps:
    
    1. Exit the current `nix-shell`:
       ```sh
       exit
       ```
    
    2. Modify your `shell.nix` file to include the new packages.
    
    3. Start a new `nix-shell` session:
       ```sh
       nix-shell
       ```
    
    This will recreate the environment with the new dependencies specified in your updated `shell.nix` file.
    
    If you want to avoid the overhead of restarting `nix-shell` frequently, consider using other development tools or strategies that can provide more dynamic package management within your development workflow. However, for `nix-shell`, restarting is necessary to pick up changes in `shell.nix`.}} \\
    \midrule
    \textbf{\scriptsize Gaussian Sample} & \raggedright\arraybackslash\scriptsize{\hl{Unfortunately, it is not possible to update the `shell.nix` file and have those changes take effect without restarting the `nix-shell`. The `nix-shell` environment is created based on the `shell.nix` file when you initially start the shell session. Any changes to `shell.nix` after the shell has been started will not be reflected until you restart the `nix-shell`.
    
    Here are the steps you would typically follow to apply changes:
    
    1. **Edit `shell.nix`:** Make the necessary changes to add new packages.
    
    2. **Exit the current `nix-shell`:** You can do this by simply typing `exit`.
    
    3. **Restart `nix-shell`:** Run `nix-shell` again to enter a new shell environment with the updated configurations.
    
    If you frequently need to test changes to your `shell.nix`, you might consider scripting these steps to make the process quicker and more convenient.}} \\
    \bottomrule
    \end{tabular}
    }
    \caption{Responses of different weighted assignment on Qwen2.5-32B-Instruct. }
    \label{tab:case_32B_fix_gaussian}
\end{table}
\hl{
Although both models provide correct answers to the current problem, the answers generated by the model after Gaussian sample are more concise and include recommendations for the user to make future predictions. Such responses are often more helpful.
}

\clearpage
\subsection{Compared to MODPO}
In these section, we provided some cases to compare {\ourtitle} with MODPO.
\begin{table}[h]
    \centering
    \resizebox{0.8\textwidth}{!}{
    \begin{tabular}{>{\centering\arraybackslash}m{2cm}>{\centering\arraybackslash}m{7cm}}
    \toprule
    \textbf{\scriptsize prompt} & \centering\arraybackslash\scriptsize{Write a code block in Markdown containing an example of a code block in Markdown. Don't forget those quadruple backticks.}\\
    \midrule
    \textbf{\scriptsize {\ourtitle}} & \raggedright\arraybackslash\scriptsize{Here is a code block in Markdown that demonstrates how to create another code block, including the quadruple backticks and the language identifier:````markdown```markdown. This is an example of a code block in Markdown. You can see that it is formatted to look like it's not part of the regular text flow.'''} \\
    \midrule
    \textbf{\scriptsize MODPO} & \raggedright\arraybackslash\scriptsize{```markdown``` Example of a code block in Markdown, print("Hello, world!")```} \\
    \midrule
    \textbf{\scriptsize Judgment} & \raggedright\arraybackslash\scriptsize{Compared to MODPO, {[\textbf{\ourtitle}]} provides a complete example that illustrates how to create a nested code block in Markdown using quadruple backticks} \\
    \bottomrule
    \end{tabular}
    }
    \caption{Responses of MODPO and {\ourtitle} with same prompt.}
    \label{tab:case_modpo_amopo_1}
\end{table}
\begin{table}[h]
    \centering
    \resizebox{0.8\textwidth}{!}{
    \begin{tabular}{>{\centering\arraybackslash}m{2cm}>{\centering\arraybackslash}m{7cm}}
    \toprule
    \textbf{\scriptsize prompt} & \centering\arraybackslash\scriptsize{Invent a new word based on a description of the word.  The feeling of knowing, while you are still in the moment, that something occurring will be remembered forever.}\\
    \midrule
    \textbf{\scriptsize {\ourtitle}} & \raggedright\arraybackslash\scriptsize{\textbf{Momentrecolle} A combination of "moment" and "mémorécolle" (from the French "mémoriser," to memorize, and "coller," to stick), *momentrecolle* describes the profound awareness that the experience or moment you are living will be indelibly etched in your memory forever.} \\
    \midrule
    \textbf{\scriptsize MODPO} & \raggedright\arraybackslash\scriptsize{\textbf{Momenticent} A blend of "moment" and "centennial," *momenticent* captures the profound awareness that an experience or event being lived through will be etched indelibly in one\'s memory, potentially becoming a milestone that defines or significantly influences one\'s life for years to come.} \\
    \midrule
    \textbf{\scriptsize Judgment} & \raggedright\arraybackslash\scriptsize{\textbf{{\ourtitle}} combines aspects of both the word 'moment' and a thoughtful incorporation of French words that emphasize memory retention, making it a unique and memorable new word that captures the essence of the described feeling. {MODPO}, while also creative, uses 'centennial' which might imply a century or 100 years, potentially confusing its intended meaning. In summary, \textbf{[{\ourtitle}]}} response is better than MODPO. \\
    \bottomrule
    \end{tabular}
    }
    \caption{Responses of MODPO and {\ourtitle} with same prompt.}
    \label{tab:case_modpo_amopo_2}
\end{table}

\begin{table}[h]
    \centering
    \resizebox{0.8\textwidth}{!}{
    \begin{tabular}{>{\centering\arraybackslash}m{2cm}>{\centering\arraybackslash}m{7cm}}
    \toprule
    \textbf{\scriptsize prompt} & \centering\arraybackslash\scriptsize{write an introduction of a person for resume who worked as an in-house IT for 18 years, managed team of 4 and 9 site and satellite offices with total 200 users. He's familiar with network and system infrastructure, server virtualization, cloud services and the migration.}\\
    \midrule
    \textbf{\scriptsize {\ourtitle}} & \raggedright\arraybackslash\scriptsize{\textbf{Professional Summary:} Dedicated and experienced IT professional with 18 years of hands-on experience in managing comprehensive IT operations for a diverse organization. Successfully led a team of 4 IT specialists, overseeing network and system infrastructure for 9 site and satellite offices, supporting a total of 200 users. Expert in designing, implementing, and maintaining robust network solutions, server virtualization, and cloud services. Proven track record in facilitating seamless migrations to cloud environments, ensuring high levels of system reliability and security. Committed to staying current with the latest IT trends and technologies to drive operational efficiency and innovation.} \\
    \midrule
    \textbf{\scriptsize MODPO} & \raggedright\arraybackslash\scriptsize{\textbf{Professional Summary} Dedicated and experienced IT professional with 18 years of hands-on experience in managing in-house IT operations for a diverse organization. Proven track record of leading a team of 4 IT professionals and overseeing the IT infrastructure for 9 site and satellite offices, serving a total of 200 users. Expert in network and system infrastructure, server virtualization, cloud services, and seamless migration processes. Committed to delivering high-quality IT solutions that drive business efficiency and innovation.} \\
    \midrule
    \textbf{\scriptsize Judgment} & \raggedright\arraybackslash\scriptsize{Compared to MODPO, {[\textbf{\ourtitle}]} provides a more detailed overview of the candidate's experience and skills, highlighting not only the management of IT operations and teams but also the specific expertise in network solutions, server virtualization, and cloud services. It also mentions staying current with IT trends, which adds to the candidate's credibility and commitment to professional growth. These details can better capture the attention of potential employers.} \\
    \bottomrule
    \end{tabular}
    }
    \caption{Responses of MODPO and {\ourtitle} with same prompt.}
    \label{tab:case_modpo_amopo_3}
\end{table}
\clearpage
\subsection{Compared to CPO}

\hl{In this section, we provided some cases to compare {~\ourtitle} with CPO.}
\begin{table}[h]
    \centering
    \resizebox{0.8\textwidth}{!}{
    \begin{tabular}{>{\centering\arraybackslash}m{2cm}>{\centering\arraybackslash}m{7cm}}
    \toprule
    \textbf{\scriptsize prompt} & \centering\arraybackslash\scriptsize{\hl{Give the provided brand a motto to use as a part of its marketing strategies.

    The Ice cream shop is a small, quaint shop located in the heart of the historic district. They offer a variety of ice cream flavors, as well as Italian ice, gelato, and sorbet. The shop has a few tables and chairs, but most of their business is to-go. They offer a loyalty program where you can get a free scoop of ice cream after 10 purchases.}}\\
    \midrule
    \textbf{\scriptsize {\ourtitle}} & \raggedright\arraybackslash\scriptsize{\hl{Certainly! Here’s a charming motto that captures the essence of your ice cream shop:

    \textbf{"Indulge in History, Savor the Moment — One Scoop at a Time"}
    
    This motto highlights the nostalgic and delightful experience your shop offers, blending the historic charm of its location with the joy of delicious treats.
    }} \\
    \midrule
    \textbf{\scriptsize CPO} & \raggedright\arraybackslash\scriptsize{\hl{Brand Motto Suggestion: \textbf{"Indulge in History, One Scoop at a Time"}

    This motto encapsulates the charm and authenticity of the historic district setting while highlighting the variety and quality of their ice cream offerings. It also invites customers to return frequently to enjoy the delightful experience, thereby encouraging participation in the loyalty program.}} \\
    \midrule
    \textbf{\scriptsize Judgment} & \raggedright\arraybackslash\scriptsize{\hl{Compared to CPO, \textbf{{\ourtitle}} provides a more complete and evocative motto by incorporating the phrase \textbf{'Savor the Moment' which adds a touch of nostalgia and enjoyment.} This better captures the quaint and historic essence of the shop and creates a more appealing and memorable marketing message.}} \\
    \bottomrule
    \end{tabular}
    }
    \caption{Responses of CPO and {\ourtitle} with same prompt.}
    \label{tab:case_cpo_amopo_1}
\end{table}

\hl{Compared with CPO, we observe that the responses of {\ourtitle} is more human-like and more enjoyment.}
\begin{table}[h]
    \centering
    \resizebox{0.8\textwidth}{!}{
    \begin{tabular}{>{\centering\arraybackslash}m{2cm}>{\centering\arraybackslash}m{7cm}}
    \toprule
    \textbf{\scriptsize prompt} & \centering\arraybackslash\scriptsize{\hl{Make the article available in a second language by translating it into it.

    Dentistry, also known as dental medicine and oral medicine, is the branch of medicine focused on the teeth, gums, and mouth. It consists of the study, diagnosis, prevention, management, and treatment of diseases, disorders, and conditions of the mouth, most commonly focused on dentition (the development and arrangement of teeth) as well as the oral mucosa. Dentistry may also encompass other aspects of the craniofacial complex including the temporomandibular joint. The practitioner is called a dentist.
    Translate to French:}}\\
    \midrule
    \textbf{\scriptsize {\ourtitle}} & \raggedright\arraybackslash\scriptsize{\hl{La dentisterie, également connue sous les noms de médecine dentaire et de médecine orale, est la branche de la médecine qui se concentre sur les dents, les gencives et la bouche. Elle englobe l'étude, le diagnostic, la prévention, la gestion et le traitement des maladies, troubles et conditions de la bouche, en se concentrant le plus souvent sur la dentition (le développement et l'arrangement des dents) ainsi que la muqueuse buccale. La dentisterie peut également englober d'autres aspects du complexe cranio-facial, y compris l'articulation temporo-mandibulaire. Le praticien est appelé dentiste.
    }} \\
    \midrule
    \textbf{\scriptsize CPO} & \raggedright\arraybackslash\scriptsize{\hl{La dentisterie, aussi appelée médecine dentaire et médecine orale, est une branche de la médecine qui se concentre sur les dents, les gencives et la bouche. Elle inclut l'étude, le diagnostic, la prévention, la gestion et le traitement des maladies, troubles et conditions de la bouche, principalement axée sur la dentition (le développement et l'arrangement des dents) ainsi que la muqueuse buccale. La dentisterie peut également englober d'autres aspects de la structure crânio-faciale, notamment la jointure temporomandibulaire. Le praticien est appelé dentist. 

    Note: The French term for the practitioner is actually "dentiste" not "dental" which would mean "dental" in English. I maintained "dentist" in the last sentence to stay consistent with the original sentence structure you provided.
    }} \\
    \midrule
    \textbf{\scriptsize Judgment} & \raggedright\arraybackslash\scriptsize{\hl{The preferred answer is {\ourtitle}, \textbf{{\ourtitle}} provides a correct and accurate translation of the given text from English to French. It maintains the original meaning and technical terminology, making it more appropriate for conveying precise medical information in French. CPO, although close, has minor inconsistencies and unnecessary explanations that detract from the straightforward nature of the request.}} \\
    \bottomrule
    \end{tabular}
    }
    \caption{Responses of CPO and {\ourtitle} with same prompt.}
    \label{tab:case_cpo_amopo_2}
\end{table}

\clearpage
\subsection{Dimensions case study}
\label{different_multi_object}
\begin{table}[h]
    \centering
    \resizebox{0.7\textwidth}{!}{
    \begin{tabular}{>{\centering\arraybackslash}m{2cm}>{\centering\arraybackslash}m{7cm}}
    \toprule
    \textbf{\scriptsize prompt} & \centering\arraybackslash\scriptsize{\hl{I want to talk about "Smooth",  a song by Rob Thomas.}}\\
    \midrule
    \textbf{\scriptsize {\ourtitle}} & \raggedright\arraybackslash\scriptsize{\hl{"Smooth" is actually a song made famous by Santana featuring Rob Thomas, not solely by Rob Thomas. Released in 1999 as part of the album "Supernatural," the song became a massive hit, blending Latin rhythms with rock and pop elements. It features Rob Thomas's vocals and songwriting, with Santana providing the distinctive guitar work and Latin music influence.

    The song reached number one on the US Billboard Hot 100 and won several Grammy Awards, including Record of the Year and Song of the Year in 2000. It's known for its catchy melody, smooth guitar riffs, and Thomas's soulful vocals. If you have any specific questions about the song or would like to discuss certain aspects of it, feel free to ask!}} \\
    \midrule
    \textbf{\scriptsize CPO} & \raggedright\arraybackslash\scriptsize{\hl{"Smooth" is a popular song written by Rob Thomas, best known as the lead singer of the rock band Matchbox Twenty, and it was a big hit when it was released as a collaborative single between Rob Thomas and latin-guitarist Santana in 1999. The song appears on Santana's album "Supernatural." 

    The song earned critical acclaim and commercial success, topping the charts in many countries and becoming one of the best-selling singles of all time. It won three Grammy Awards, including "Record of the Year," in 2000. 
    
    The song features Thomas's smooth vocals combined with Santana's distinctive guitar work. The lyrics tell a story of a man who speaks to a woman in Spanish but doesn't know the language, yet he seems to be able to charm her and convince her to leave with him. The song's hook features the line "I can talk a good game," which refers to the narrator's ability to charm and seemingly speak the language fluently.
    
    "Smooth" is known for its catchy melody, memorable guitar solos, and unique fusion of rock and Latin music. The song's success helped to introduce Latin music to a broader audience and has remained a popular classic since its release. 
    
    Is there anything in particular about the song or its impact that you would like to discuss?}} \\
    \midrule
    \textbf{\scriptsize Helpfulness Judgment} & \raggedright\arraybackslash\scriptsize{\hl{\textbf{{\ourtitle}} is more helpful in this case as it directly addresses the user's instruction by providing essential facts about the song 'Smooth,' focusing on the key collaboration between Santana and Rob Thomas, its release details, and the song's success. \textbf{It avoids unnecessary elaboration and stays concise}, which might be preferable if the user was looking for clear and accurate information without getting into deeper analysis.}} \\
    \midrule
    \textbf{\scriptsize Correctness Judgment} & \raggedright\arraybackslash\scriptsize{\hl{\textbf{\ourtitle} is more correct \textbf{in terms of the factual details about the collaboration, chart success, and Grammy wins, ensuring better correctness} than \textbf{MODPO}, \textbf{which introduces inaccurate interpretations of the song's lyrics.}}} \\
    \midrule
    \textbf{\scriptsize Instruct Follow Judgment} & \raggedright\arraybackslash\scriptsize{\hl{\textbf{\ourtitle} \textbf{follows the instruction more closely by focusing on the song's key elements: collaboration with Santana, its release, and major achievements like chart success and Grammy Awards}. \textbf{MODPO}, while informative, diverges slightly \textbf{by adding speculative interpretations of the lyrics and the song’s broader impact on Latin music,} which goes beyond the initial request to simply talk about the song.}} \\
    \bottomrule
    \end{tabular}
    }
    \caption{Responses of CPO and {\ourtitle} with same prompt.}
    \label{tab:case_cpo_amopo_1}
\end{table}

\begin{table}[h]
    \centering
    \resizebox{0.8\textwidth}{!}{
    \begin{tabular}{>{\centering\arraybackslash}m{2cm}>{\centering\arraybackslash}m{7cm}}
    \toprule
    \textbf{\scriptsize prompt} & \centering\arraybackslash\scriptsize{\hl{Find the directions between the given locations.

    From: Rowan Avenue Elementary School, 600 S Rowan Ave, Los Angeles, CA 90023, United States
    
    To: Calvary Cemetery and Mortuary, 4201 Whittier Blvd, Los Angeles, CA 90023, United States}}\\
    \midrule
    \textbf{\scriptsize {\ourtitle}} & \raggedright\arraybackslash\scriptsize{\hl{To find directions from Rowan Avenue Elementary School (600 S Rowan Ave, Los Angeles, CA 90023) to Calvary Cemetery and Mortuary (4201 Whittier Blvd, Los Angeles, CA 90023), you can follow these general steps:

    1. Head west on S Rowan Ave toward E 62nd St for approximately 0.1 miles.
    
    2. Turn right onto S Eastern Ave and continue for about 1.4 miles.
    
    3. Take a left turn to merge onto Whittier Blvd.
    
    4. Continue on Whittier Blvd for approximately 1.1 miles.
    
    5. Calvary Cemetery and Mortuary will be on your right.
    
    For the most accurate and up-to-date directions, it is recommended to use a GPS device or a mapping service like Google Maps, Apple Maps, or Waze, as road conditions and construction can change.}} \\
    \midrule
    \textbf{\scriptsize MODPO} & \raggedright\arraybackslash\scriptsize{\hl{To travel from Rowan Avenue Elementary School to Calvary Cemetery and Mortuary, you can follow these directions:

    1. Start at Rowan Avenue Elementary School, located at 600 S Rowan Ave, Los Angeles, CA 90023.
    
    2. Head west on S Rowan Ave toward E 41st St.
    
    3. Turn left onto E 41st St.
    
    4. Continue straight onto E 41st St, which will turn into S Gramercy Pl.
    
    5. Turn right onto E 43rd St.
    
    6. Turn left onto S Gramercy Pl.
    
    7. Turn right onto S San Pedro St.
    
    8. Continue on S San Pedro St, which will turn into Whittier Blvd.
    
    9. Calvary Cemetery and Mortuary will be on your right at 4201 Whittier Blvd, Los Angeles, CA 90023.
    
    Alternatively, you can use a navigation app like Google Maps or Waze to get the most accurate and up-to-date directions.}} \\
    \midrule
    \textbf{\scriptsize Helpfulness Judgment} & \raggedright\arraybackslash\scriptsize{\hl{\textbf{{\ourtitle}} is more helpful as it provides a clear and direct set of directions with specific landmarks and turns between Rowan Avenue Elementary School and Calvary Cemetery and Mortuary. \textbf{It also includes a helpful note recommending the use of a GPS or mapping service for the most accurate and up-to-date information, which is practical advice.} On the other hand, \textbf{MODPO} provides a more complicated route \textbf{with unnecessary detours and multiple streets} that are not as straightforward as the path suggested in {\ourtitle}. The additional routes in MODPO make it less practical for someone seeking simple and clear directions. Therefore, {\ourtitle} is more aligned with the user's need for helpful, direct, and useful information.}} \\
    \midrule
    \textbf{\scriptsize Correctness Judgment} & \raggedright\arraybackslash\scriptsize{\hl{\textbf{{\ourtitle}} \textbf{provides more accurate and concise directions, with specific distances and clear turn-by-turn instructions from Rowan Avenue Elementary School to Calvary Cemetery and Mortuary.} The directions follow a straightforward route, making it easier to navigate. It avoids unnecessary detours and accurately reflects the location of both destinations. \textbf{MODPO}, while containing some correct directions, \textbf{introduces unnecessary complexity and detours that make it less efficient}. Therefore, {\ourtitle} is the most correct, relevant, and practical response.}} \\
    \midrule
    \textbf{\scriptsize Instruct Follow Judgment} & \raggedright\arraybackslash\scriptsize{\hl{\textbf{\ourtitle} \textbf{follows the instruction closely by providing clear, concise directions from the given locations.} It meets the task goals without unnecessary complexity, ensuring full alignment with the instruction following requirements.}} \\
    \bottomrule
    \end{tabular}
    }
    \caption{Responses of MODPO and {\ourtitle} with same prompt.}
    \label{tab:case_modpo_amopo_1}
\end{table}